%% file: ms.tex
\documentclass[letterpaper,twocolumn,10pt]{article}
\usepackage{usenix-2020-09}

\usepackage{tikz}
\usepackage{amsmath}

\usepackage{filecontents}

\usepackage{caption}

\usepackage{booktabs} %
\usepackage{enumitem}
\usepackage{balance}
\usepackage{graphicx}
\usepackage{amsmath,lipsum}  
\usepackage{fixltx2e}
\usepackage{url}
\usepackage{color}
\usepackage{listings}
\usepackage{xspace}
\usepackage{makecell}
\usepackage{fancyvrb}
\usepackage{verbatim}
\usepackage{etoolbox}
\usepackage{multirow}
\usepackage{comment}
\usepackage{hyperref}
\usepackage[normalem]{ulem}
\usepackage[noend]{algpseudocode}
\usepackage{algorithm}
\usepackage[normalem]{ulem}
\usepackage{float}
\usepackage{subcaption}

\input{defs}
 
\setlist{nosep}
\newcommand{\alignvspacetop}{-0.8em}
\newcommand{\alignvspacebottom}{-2.5em}
\newcommand{\aligntop}{\vspace{\alignvspacetop}}
\newcommand{\alignbottom}{\\[\alignvspacebottom]}

\begin{document}

\title{\sys: Derivation-based Tensor Program Optimizer}

\author{
{\rm Liyan Zheng \dag, Haojie Wang \dag, Jidong Zhai\dag, Muyan Hu\dag, Zixuan Ma\dag,} \\ {\normalfont Tuowei Wang\dag, Shizhi Tang\dag, Lei Xie\dag, Kezhao Huang\dag, Zhihao Jia \ddag}\\
\dag Tsinghua University \\
\ddag Carnegie Mellon University
} %

\maketitle

\vspace{\baselineskip}

\pagestyle{plain}

\begin{abstract}
\input{abstract}
\end{abstract}

\input{introduction}
\input{background}
\input{rules2}
\input{search}
\input{implementation}
\input{evaluation}
\input{related}
\input{conclusion}

\bibliographystyle{plain}
\bibliography{ref}

\end{document}

%% file: defs.tex
\usepackage{xcolor}
\usepackage{amsmath}
\usepackage{amssymb}
\usepackage{amsthm}
\usepackage{thmtools}
\usepackage{nameref,cleveref}

\newcommand{\cblack}{\textcolor{black}}

\definecolor{darkgreen}{rgb}{0,0.5,0}

\definecolor{purple}{rgb}{0.3,0.0,0.6}

\definecolor{auburn}{rgb}{0.43,0.21,0.1}

\definecolor{gray}{rgb}{0.7,0.7,0.7}

\definecolor{dgray}{rgb}{0.5,0.5,0.5}

\newcommand{\newcnt}[2][]{{#2}}

\definecolor{darkgreen}{rgb}{0,0.5,0}

\newcommand{\para}[1]{\noindent \textbf{#1 }}
\newcommand{\er}[1]{\mbox{\rm\em #1}}
\newcommand{\m}[1]{\mathcal{#1}}

\newcommand{\sys}{\textsc{Ollie}\xspace}
\newcommand{\stage}{scope\xspace}

\newcommand{\stages}{scopes\xspace}

\newcommand{\opL}{\mathop{\vphantom{\sum}\mathchoice
  {\vcenter{\hbox{\LARGE L}}}
  {\vcenter{\hbox{\large L}}}{\mathrm{L}}{\mathrm{L}}}\displaylimits}
\newcommand{\opLd}{\opL}
\newcommand{\sumd}{\sum}
\newcommand{\pos}[1]{[\mathbf{\tau}(#1)]}
\newcommand{\posB}[1]{[\mathbf{\pi}(#1)]}
\newcommand{\mathT}{\tensors{T}}

\newcommand{\tensors}[1]{\mathbf{#1}}

\newcommand{\itab}{iterator mapping table\xspace}
\newcommand{\Itab}{Iterator mapping table\xspace}

\newcommand{\eop}{eOperator\xspace}
\newcommand{\eops}{eOperators\xspace}

\crefformat{section}{\S#2#1#3}
\crefformat{subsection}{\S#2#1#3}
\crefrangeformat{section}{\S#2#1#3}
\crefrangeformat{subsection}{\S#2#1#3}

%% file: abstract.tex
Boosting the runtime performance of deep neural networks (DNNs) is critical due to their wide adoption in real-world tasks.
Existing approaches to optimizing the tensor algebra expression of a DNN only consider expressions {\em representable} by a fixed set of predefined operators, missing possible optimization opportunities between {\em general} expressions.
We propose \sys, the first derivation-based tensor program optimizer. \sys optimizes tensor programs by leveraging transformations between {\em general} tensor algebra expressions, enabling a significantly larger expression search space that includes those supported by prior work as special cases.
\sys uses a {\em hybrid} derivation-based optimizer that effectively combines explorative and guided derivations to quickly discover highly optimized expressions. 
Evaluation on seven DNNs shows that \sys{} can outperform existing optimizers by up to $2.73\times$ ($1.46\times$ on average) on an A100 GPU and up to $2.68\times$ ($1.51\times$) on a V100 GPU, respectively.

%% file: introduction.tex
\section{Introduction}
Fast execution of deep neural networks (DNNs) is critical in a variety of tasks, such as autonomous driving~\cite{grigorescu2020survey,liu2020computing,kiran2021deep}, object detection~\cite{he2017mask,girshick2015fast}, speech recognition~\cite{amodei2016deep,gulati2020conformer}, and machine translation~\cite{wu2016google,vaswani2017attention}.
A DNN is generally represented as a {\em tensor program}, which is a directed graph defining the tensor operators (e.g., convolution, matrix multiplication) performed on a set of tensors (i.e., $n$-dimensional arrays).

To optimize the runtime performance of a DNN, many existing frameworks, including TensorFlow, PyTorch, and TensorRT, rely heavily on {\em manually-designed} rules to map the tensor operators in the DNN to vendor-provided kernel libraries~\cite{tensorflow2015-whitepaper,pytorch, tensorrt}.
Although widely used, this rule-based approach requires extensive engineering efforts and only performs optimizations manually discovered.

Recent work has proposed a variety of {\em automated} approaches that optimize DNN computation by searching over a set of candidate program transformations.
We classify existing approaches into two categories based on their search spaces.

The first category of work explores the {\it expression} search space.
For a given tensor program, a {\em tensor algebra expression} defines a potential algorithm to compute the program's output from the input tensors.
For example, the computation for a convolution can be expressed as direct convolution, image-to-column, or other algorithms~\cite{chellapilla2006high, cudnn}.
TASO~\cite{jia2019taso} and PET~\cite{pet} explore the expression search space using automatically generated transformations between tensor operators. 
Both of them use a {\em superoptimization-based} approach that generates expression transformations by enumerating possible combinations of a fixed set of predefined operators.
While TASO and PET have demonstrated superior performance over rule-based optimizers, they only discover transformations between expressions that can be constructed using only the predefined operators. 
These expressions are referred to as {\em predefined operator representable (POR)} expressions.

The second category of work is motivated by Halide's idea of compute/schedule separation~\cite{halide} and optimizes tensor programs by exploring the {\em schedule} search space.
Examples include TVM~\cite{tvm} and Ansor~\cite{ansor}.
For a given expression, a {\em schedule} specifies a strategy to execute the computation defined by the expression on a particular hardware device.
Although effective at generating high-performance low-level kernels for individual expressions, the schedule-based optimizers only consider schedules that faithfully execute the computation defined by a given expression but miss optimizations that depend on other functionally equivalent expressions.

\begin{figure}
	\centering
	\includegraphics[width=1\linewidth]{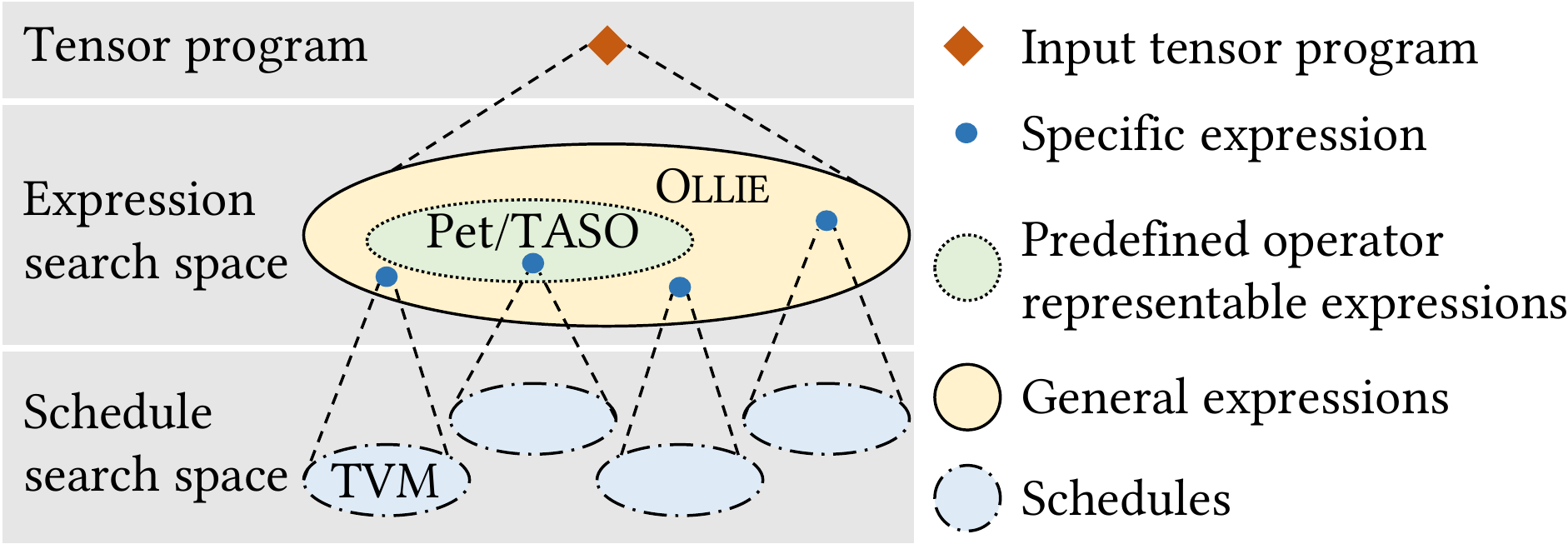}
	\vspace{-2em}
	\caption{Comparing \sys's search space with those of prior work. } 
	\label{fig:diff}
	\vspace{-2em}
\end{figure}

\paragraph{Our Approach.}
This paper presents \sys, the first {\em derivation-based} tensor program optimizer. \sys explores the {\em expression} search space and therefore is orthogonal and can be combined with existing schedule-based optimizers.
\Cref{fig:diff} depicts a comparison between \sys's search space with those of existing automated approaches.
A key difference between \sys and prior work (i.e., TASO and PET) is that \sys is able to explore a significantly larger search space of {\em general} expressions, which includes the POR expressions supported by TASO and PET as special cases. 
This generalization unlocks a new class of expression-level optimizations; for example, as we will discuss in \Cref{sec:motivation}, \sys is able to automatically discover multiple ways to transform a convolution to other (non-POR) expressions, some of which outperform the best convolution algorithms of prior work.

A challenge \sys must address is automatically discovering transformation opportunities between {\em general} expressions.
Directly considering general expressions in existing superoptimization-based approach is infeasible, since a superoptimizer discovers transformations by enumerating all possible operators, which can be infinitely many for general expressions.
To address this challenge, a key idea behind \sys is a {\em derivation-based} mechanism that automatically transforms a tensor algebra expression to functionally equivalent alternatives by applying a collection of novel derivation rules.
Most derived expressions cannot be simply decomposed into predefined operators; thus we introduce {\em \eops} to represent the non-POR parts of an expression.
\eops enable \sys to generate performance-improving transformations between {\em general} expressions.
For example, as we will discuss in \Cref{sec:motivation} and shown in \Cref{fig:conv2gemm}, \sys can optimize a convolution by transforming it to a matrix multiplication (i.e., a predefined operator) and an element-wise addition with customized offsets (an \eop).

\begin{figure}
	\centering
	\includegraphics[width=1\linewidth]{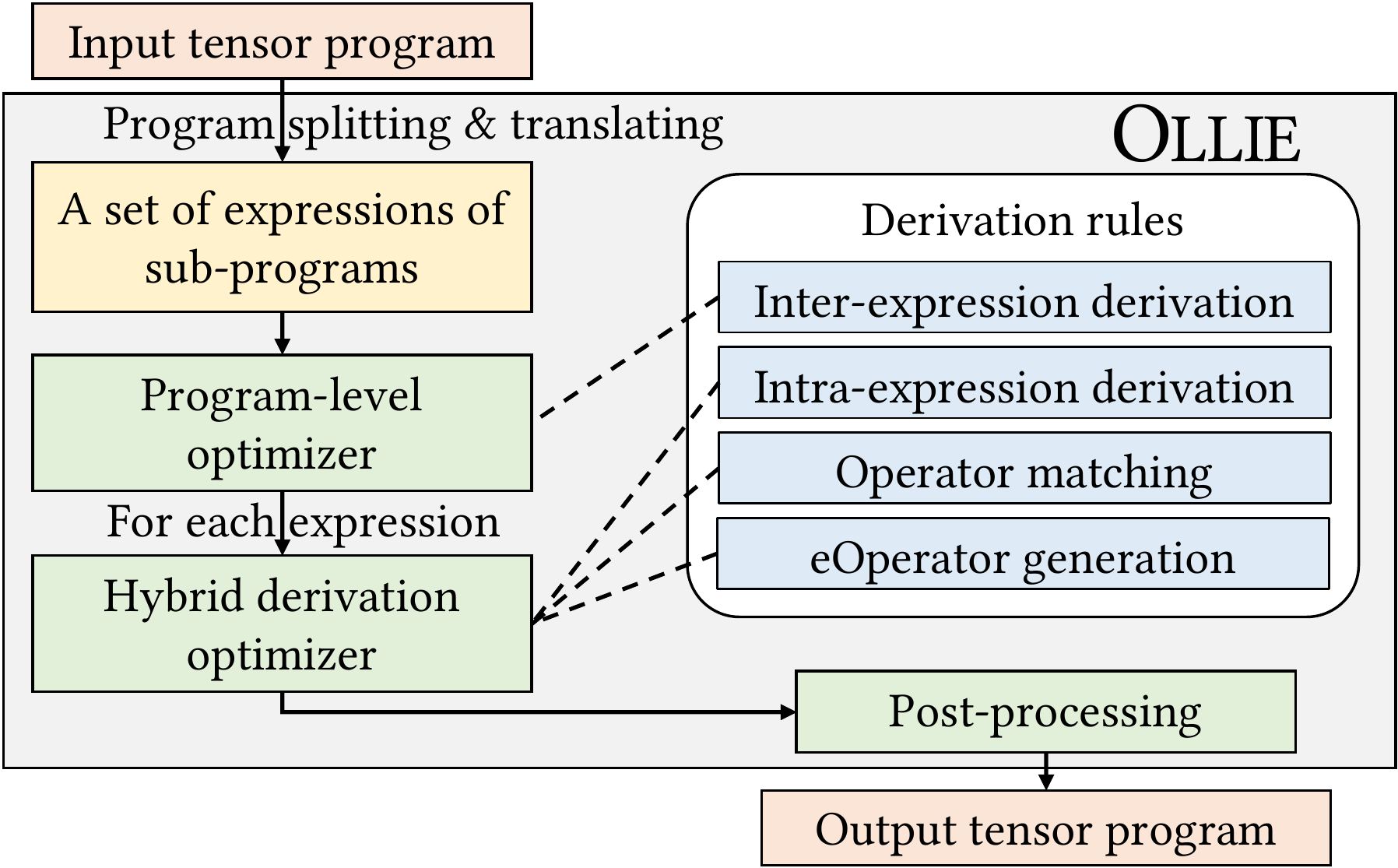}
	\vspace{-2em}
	\caption{\sys overview}
	\label{fig:overview}
	\vspace{-2em}
\end{figure} 

\Cref{fig:overview} shows an overview of \sys. For an input tensor program, \sys first splits the program into a number of subprograms by adopting the program splitting mechanism from prior work~\cite{pet, ansor}.
Each subprogram is translated to a tensor algebra expression.
Second, \sys's {\em program-level optimizer} applies inter-expression derivation rules to explore optimization opportunities across expressions, such as fusing multiple expressions into one.

A core component of \sys is a {\em hybrid derivation optimizer} that optimizes each subprogram by exploring its expression search space.
\sys uses a set of intra-expression derivation rules to transform a candidate expression to equivalent alternatives.
During the search, \sys opportunistically matches a part of the expression with predefined operators using an {\em operator matching} algorithm, which allows \sys to leverage the vendor-provided heavily-optimized kernels (e.g., cuDNN~\cite{cudnn} and cuBLAS~\cite{cublas}) to execute the matched part of the expression.
\sys generates \eops for the remaining part of the expression.

Optimizing an expression may require applying a long sequence of derivation rules (e.g., our motivating example in \Cref{sec:motivation} uses seven derivation rules), which cannot be efficiently discovered by a fully randomized search algorithm.
To address this challenge, \sys employs a {\em hybrid} search approach that performs {\em explorative} derivations for the initial rounds of the search, allowing \sys to explore a comprehensive collection of expressions. For each candidate expression, \sys then applies {\em guided} derivations to derive the expression toward target tensor operators using selected rules.
This hybrid approach allows \sys to discover complex optimizations, each of which requires a long sequence of derivations, under a reasonable search budget. 

Finally, \sys selects the best discovered expression for each subprogram and post-processes the expressions to construct an optimized tensor program.

We evaluate \sys on seven real-world DNN models that cover a variety of machine learning tasks.
We compare \sys with state-of-the-art frameworks on two widely used GPU platforms, NVIDIA Tesla A100 and V100.
Evaluation shows that \sys is up to $2.73\times$ faster than existing tensor program optimizers.
Even for heavily optimized DNNs, such as ResNet-18, \sys can still achieve a $1.33\times$ speedup.
The significant performance improvement indicates that \sys benefits from the new optimization opportunities enabled by derivation-based expression-level optimizations.

This paper makes the following contributions:

\begin{itemize}[leftmargin=*]
    \item We extend the expression search space from predefined-operator-representable expressions to general expressions.
    \item We present the first attempt to explore the significantly larger expression search space using a derivation-based mechanism.
    \item We build \sys, an implementation of the above techniques with over 23K lines of C++ code, which achieves up to $2.73\times$ speedup over existing tensor program optimizers.
\end{itemize}

%% file: background.tex
\section{Motivating Example}
\label{sec:motivation}
\begin{figure*}[h!]
    \centering
    \begin{subfigure}[t]{0.42\textwidth}
        \centering
        \includegraphics[height=15em,keepaspectratio]{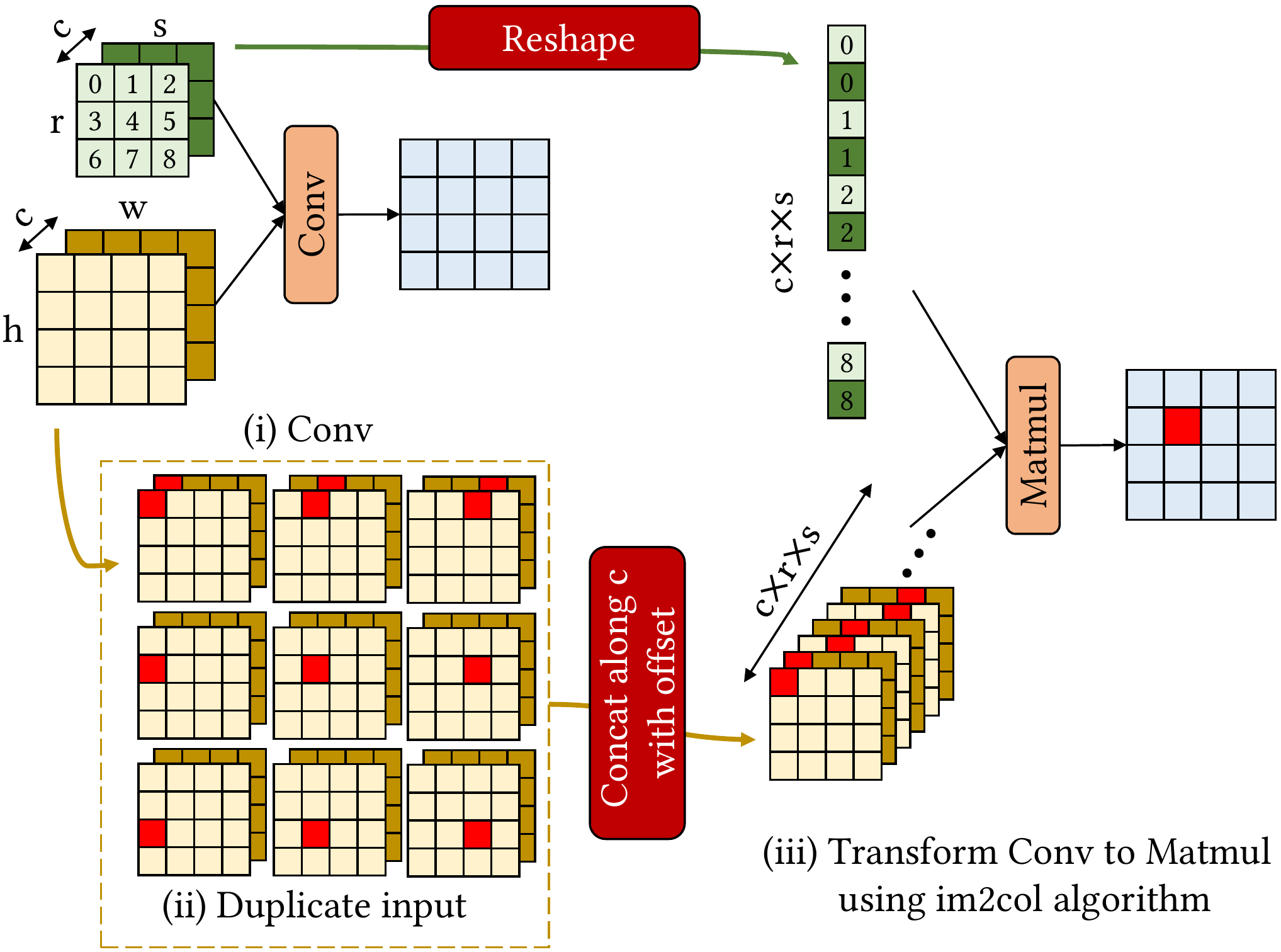}
        \caption{Image-to-column optimization}
        \label{fig:img2col}
    \end{subfigure}\quad
    \begin{subfigure}[t]{0.5\textwidth}
        \centering
        \includegraphics[height=15em,keepaspectratio]{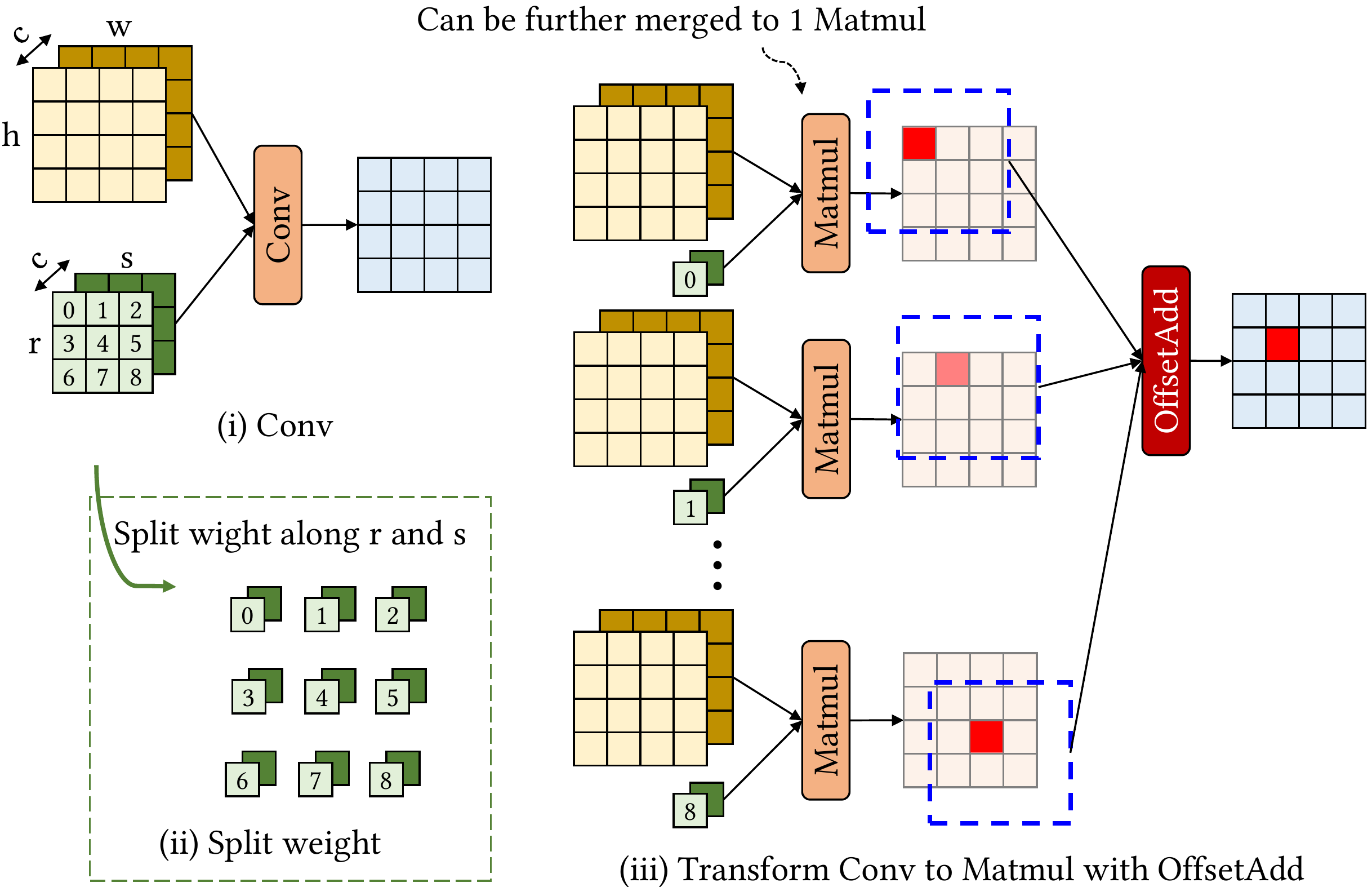}
        \caption{Optimization found by \sys}
        \label{fig:conv2gemm}
    \end{subfigure}
    \vspace{-1em}
    \caption{Transform {\tt Conv} computation to {\tt Matmul} computation. \eops are represented as red rounded rectangles. Both figure(i) show a simplified convolution omitting batch dimension and channel dimension of output.
    It takes an input tensor with size $(c, h, w)$, where $c$, $h$, and $w$ represent the input channels, height, and width of images, and a weight tensor with the size of $(c, r, s)$, where $r$ and $s$ are the height and width of the convolution kernel. }
    \label{fig:motivation}
    \vspace{-1em}
\end{figure*}

As a motivating example, \Cref{fig:img2col} demonstrates the image-to-column optimization~\cite{vasudevan2017parallel}, which transforms a $3\times 3$ convolution (\Cref{fig:img2col}(i)) to a matrix multiplication for devices supporting efficient matrix computations, such as TPUs and GPUs with tensor cores~\cite{jouppi2017datacenter, tensorcore}.
As shown in \Cref{fig:img2col}(ii), the first step of the transformation is duplicating an input image into 9 identical replicas.
Second, 
we concatenate the 9 images into one using different offsets, such that pixels highlighted in red are grouped together in \Cref{fig:img2col}(iii).
Finally, we change the layout of the weight (shown in green) to match the dimension of the concatenated image and perform a matrix multiplication, which yields an identical output as the original convolution operator. 
This optimization has been widely used in different DNN libraries and is implemented by human experts, but \sys can find and implement it automatically using expression derivation.

\sys is able to find more optimizations.
For the same convolution in \Cref{fig:conv2gemm}(i), \sys first splits the weight tensor into $9$ small weight tensors along $r$ and $s$ in \Cref{fig:conv2gemm}(ii). 
Then, in \Cref{fig:conv2gemm}(iii), each small weight tensor performs matrix multiplication with the input tensor. 
\sys further merges these matrix multiplications into a single one.
Finally, the intermediate results are added together to generate an output tensor.
Since the addition is taken on each blue dashed region of intermediate tensors, the addition is a customized \eop\ {\tt OffsetAdd}, which is automatically generated by \sys{}.
This transformation can further improve the performance of evaluated convolution operators by more than two times compared with the cuDNN libarary on Tesla A100. 

However, previous frameworks cannot generated such transformations since:
1) The transformations require \eops, e.g., adding intermediate tensors with offset, which is outside of the POR expresssion space explored by superoptimization-based frameworks like TASO and PET.
2) The transformations change the expressions instead of the schedule, and thus cannot be generated by schedule-based optimizers like TVM or Ansor.

%% file: rules2.tex
\section{Tensor Algebra Expression}
\begin{figure}[t]
    \centering
    \includegraphics[width=\linewidth]{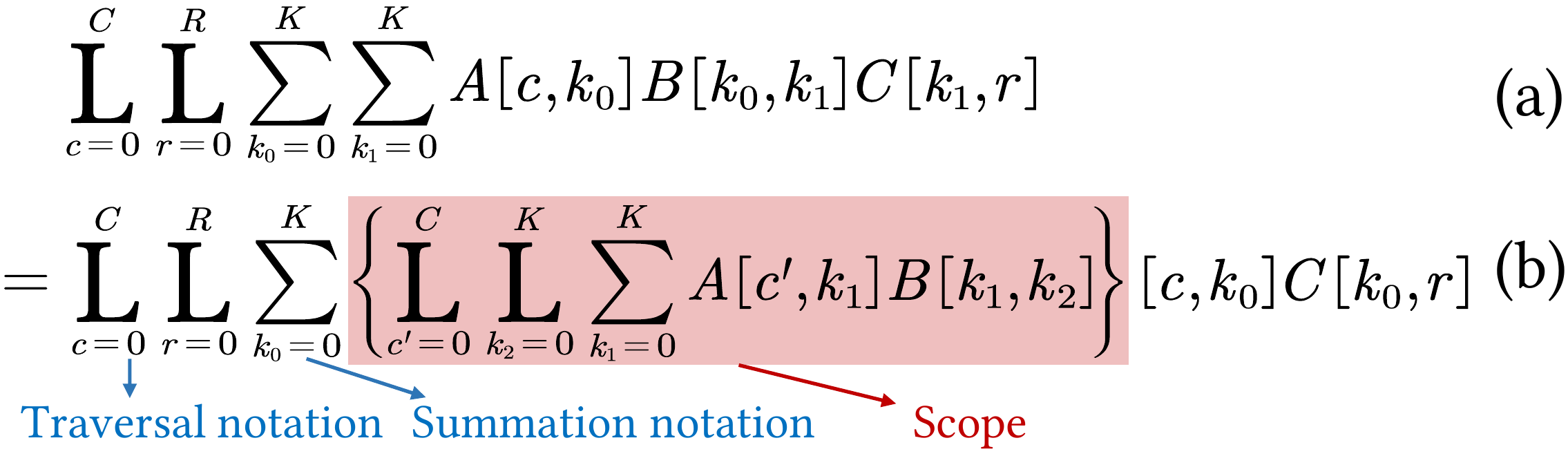}
    \vspace{-2em}
    \caption{Example tensor algebra expressions for two matrix multiplications $A\times B\times C$. The red box highlights a \stage that instantiates the intermediate result of $A\times B$.}
    \label{fig:expr}
    \vspace{-1.5em}
\end{figure}

\sys represents a tensor program as a {\em tensor algebra expression} that defines how to compute each element of the output tensor from input tensors.
For example, \Cref{fig:expr} shows the expression for multiplying three matrices. We now explain the components of an expression.

\para{Traversal and summation notations.} A {\em traversal notation}, denoted as $\opLd_{x=x_0}^{x_1}$, consists of an {\em iterator} $x$ and an {\em iterating space} $[x_0, x_1)$.
The traversal notation corresponds to a dimension of the output tensor, where the iterating space is the range of the dimension. 
The order between different traversal notations indicates the layout of the output tensor.
E.g., in \Cref{fig:expr}, $\opLd_{c=0}^{C}$ followed by $\opLd_{r=0}^{R}$ shows that the expression's output is a 2-dimensional tensor with a shape $C\times R$.

A {\em summation notation}, denoted as
$\sumd_{x=x_0}^{x_1}$, corresponds to a dimension with range $[x_0, x_1)$ that is iterated over when computing each output element. In \Cref{fig:expr}, the summation notation $\sumd_{k_0=0}^{K}$ indicates that the computation for each output element iterates over an interval with $K$ elements.
Note that \sys's expression notation is invariant under permutation of summation notations, which corresponds to different {\em schedules} of an expression and therefore is not considered in the expression search space.

Tensors are indexed by a {\em linear combination} of multiple iterators 
or the division or remainder of iterators (e.g., $h\%2$).
For simplicity, we may merge multiple iterators into one iterator vector in the paper. 
Iterator space also can be denoted by an integer set or omitted in expressions. 
For example, $\opLd_{c=0}^{C}\opLd_{r=0}^{R}$ can be represented as $\opLd_{cr}$ or $\opLd^{\mathbb{X}}_{\vec{x}}$, where $\vec{x}=(c,r)$ and $\mathbb{X}=\mathbb{C}\times\mathbb{R}$ is the iterating space. 

\para{Scope.} For a tensor program with multiple operators (e.g., two consecutive matrix multiplications $A\times B \times C$), a common optimization is to instantiate and reuse intermediate results (e.g., caching the output of $A\times B$), which avoids repetitive computation for these results. \sys introduces \stages to represent the instantiation of intermediate results and enable their reuse. Formally, a tensor algebra expression is a {\em \stage}, denoted by a surrounding $\big\{\big\}$, if the output of the expression is instantiated into a tensor, which allows subsequent computation to refer to its output as a tensor and therefore avoids repeated computation of its output.
In \Cref{fig:expr}(b), the expression corresponding to $A\times B$ becomes a \stage, which allows subsequent computation to directly refer to the output of this expression as a tensor.

Most of \sys's derivation rules (described in \Cref{sec:rules}) are based on transformations between \stages, including generating new \stages from existing ones, transforming a \stage to another form, and merging multiple \stages into one.
Transformations between scopes are essential to \sys's performance optimizations. 

\para{Padding}
Tensors may have paddings.
For example, convolution may access input tensors at a position outside of $A$'s region.
We call these positions the paddings of $A$, whose elements are $0$ if not specified.

\para{General format.}
We represent arbitrary 1-\stage expressions in the following format:
\vspace{\alignvspacetop}
\begin{align*}
    \opL_{\vec{x}}^{\mathbb{X}}\sum_{\vec{y}}^{\mathbb{Y}}f(\mathT\pos{\vec{x},\vec{y}})
    \\[\alignvspacebottom]
\end{align*}
where $\mathT=\{T_0, T_1, ...\}$ is a list of input tensors, $\tau(\vec{x},\vec{y})$ is the indexing function that computes an element index for $\mathT$ using iterators $\vec{x}$ and $\vec{y}$, and $f$ is the computation taking on the indexed elements of $\mathT$.

\section{Derivation Rules}
\label{sec:rules}

\begin{table*}[h]
\centering
\caption{Derivation rules for tensor algebra expressions}
\label{tab:derivation-rules}
\vspace{-0.5em}
\small
\begin{tabular}{cll}
\toprule
& Rules & Descriptions \\ \midrule

\multirow{3}{*}{\shortstack[m]{Inter-expression \\ derivation}}  & Expression splitting & Split an expression into multiple independent expressions \\ 
& Expression merging & Merge multiple independent expressions into one \\ 
& Expression fusion & Fusing multiple dependent expressions into one \\ 
\midrule

\multirow{5}{*}{\shortstack[m]{Intra-expression \\ derivation}} 
& Summation splitting & Split summation from one \stages into two  \\ %
& Variable substitution & Replace iterators of traversal notations with new ones \\ %
& Traversal merging & Merge two \stages into one by merging traversals  \\ %
& Boundary relaxing & Relax the range of iterators \\ 
& Boundary tightening  & Tighten  the range of iterators \\ 
\midrule

\multirow{2}{*}{\shortstack[m]{Expression \\ instantiation}} 
& Operator matching & Match a \stage with operators and replace it by a tensor \\
& \eop generation & Generate an \eop for a scope and replace it by a tensor \\ 
\bottomrule
\end{tabular}%
\vspace{-1em}
\end{table*}

To discover highly-optimized expressions for an input tensor program, \sys uses {\em derivation rules} to apply transformations on an input expression.
\Cref{tab:derivation-rules} summarizes the derivation rules used by \sys, which are divided into three categories.
The proof of these rules are provided in Appendix.

\subsection{Inter-Expression Derivation}
\label{sec:inter-expr-rule}

\para{Expression splitting} divides an expression into two independent expressions by partitioning the traversal space of the expression $\mathbb{X}$ into two subspaces $\mathbb{X}_1$ and $\mathbb{X}_2$ such that $\mathbb{X} \subseteq \mathbb{X}_1 \cup \mathbb{X}_2$. The expression splitting rule is as follows: 
\vspace{\alignvspacetop}
\begin{align*}
\opL_{\vec{x}}^{\mathbb{X}}f(\mathT\pos{\vec{x}}) \Longrightarrow 
\opL_{\vec{x}}^{\mathbb{X}_1}f(\mathT\pos{\vec{x}}) \sim \opL_{\vec{x}}^{\mathbb{X}_2}f(\mathT\pos{\vec{x}}) 
\\[\alignvspacebottom]
\end{align*}
where $\sim$ denotes the independence of the two expressions. 

\para{Expression merging} combines two independent expressions that have identical computation into a single expression by merging the traversal spaces of the original expressions:
\vspace{\alignvspacetop}
\begin{align*}
\opL_{\vec{x}}^{\mathbb{X}_1}f(\mathT\pos{\vec{x}}) \sim \opL_{\vec{x}}^{\mathbb{X}_2}f(\mathT\pos{\vec{x}}) \Longrightarrow \opL_{\vec{x}}^{\mathbb{X}}f(\mathT\pos{\vec{x}})
\\[\alignvspacebottom]
\end{align*}
where $\mathbb{X}_1 \cup \mathbb{X}_2 \subseteq \mathbb{X}$. Expression merging is the symmetric transformation of expression splitting.

\begin{figure}[]
    \centering
    \includegraphics[width=\linewidth,keepaspectratio]{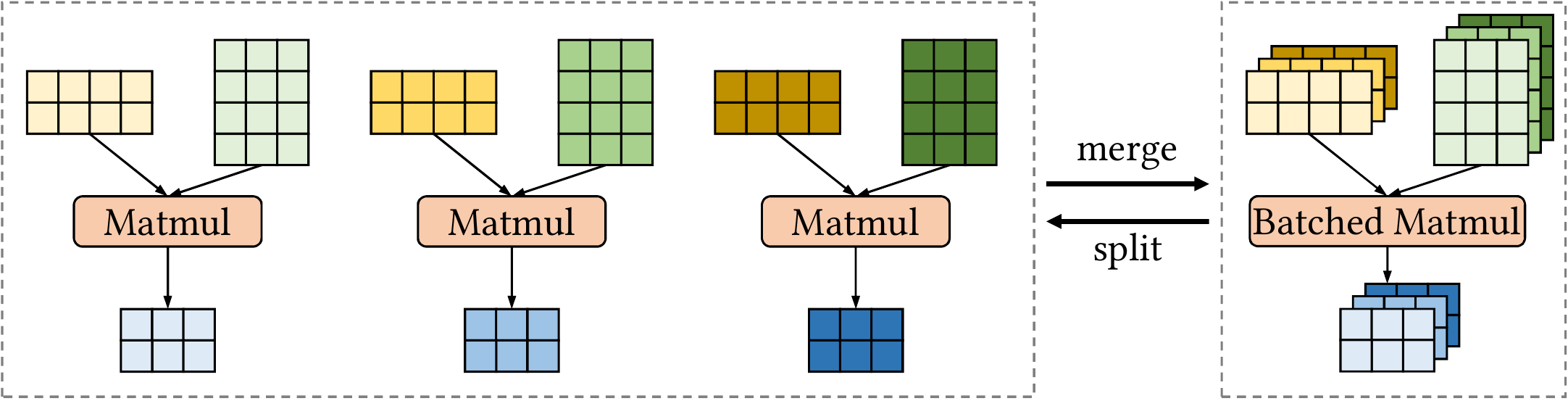}
    \vspace{-2em}
    \caption{Splitting and merging matrix multiplications.}
    \label{fig:batchgemm}
    \vspace{-2em}
\end{figure}

A typical example of expression splitting and merging is the transformation between {\tt Matmul} and {\tt BatchMatmul}: multiple matrix multiplications can be merged together to a single batch matrix multiplication, and a batch matrix multiplication can be split into multiple matrix multiplications, as shown in \Cref{fig:batchgemm}.

\para{Expression fusion} fuses multiple dependent expressions into one using the chain rule:
\aligntop
\begin{align*}
\opL_{\vec{y}}^{\mathbb{Y}}g(\tensors{T'}\posB{\vec{y}}) \circ
\opL_{\vec{x}}^{\mathbb{X}}f(\mathT\pos{\vec{x}}) \Longrightarrow
\opL_{\vec{y}}^{\mathbb{Y}}g(\{\opL_{\vec{x}}^{\mathbb{X}}f(\mathT\pos{\vec{x}})\}\posB{\vec{y}})
\alignbottom
\end{align*}
where $\m{E}_1\circ\m{E}_2$ denotes that the result of  expression $\m{E}_2$ is fed as inputs to expression $\m{E}_1$ (i.e., $\tensors{T'}$ equals to the computation result of $\opL_{\vec{x}}^{\mathbb{X}}f(\mathT\pos{\vec{x}})$).

Expression fusion allows \sys{} to consider {\em operator fusion}, a technique that fuses the computation of multiple operators into a single kernel to reduce data movement and kernel launch overhead.

\begin{figure*}[h!]
    \centering
    \includegraphics[width=0.95\linewidth,keepaspectratio]{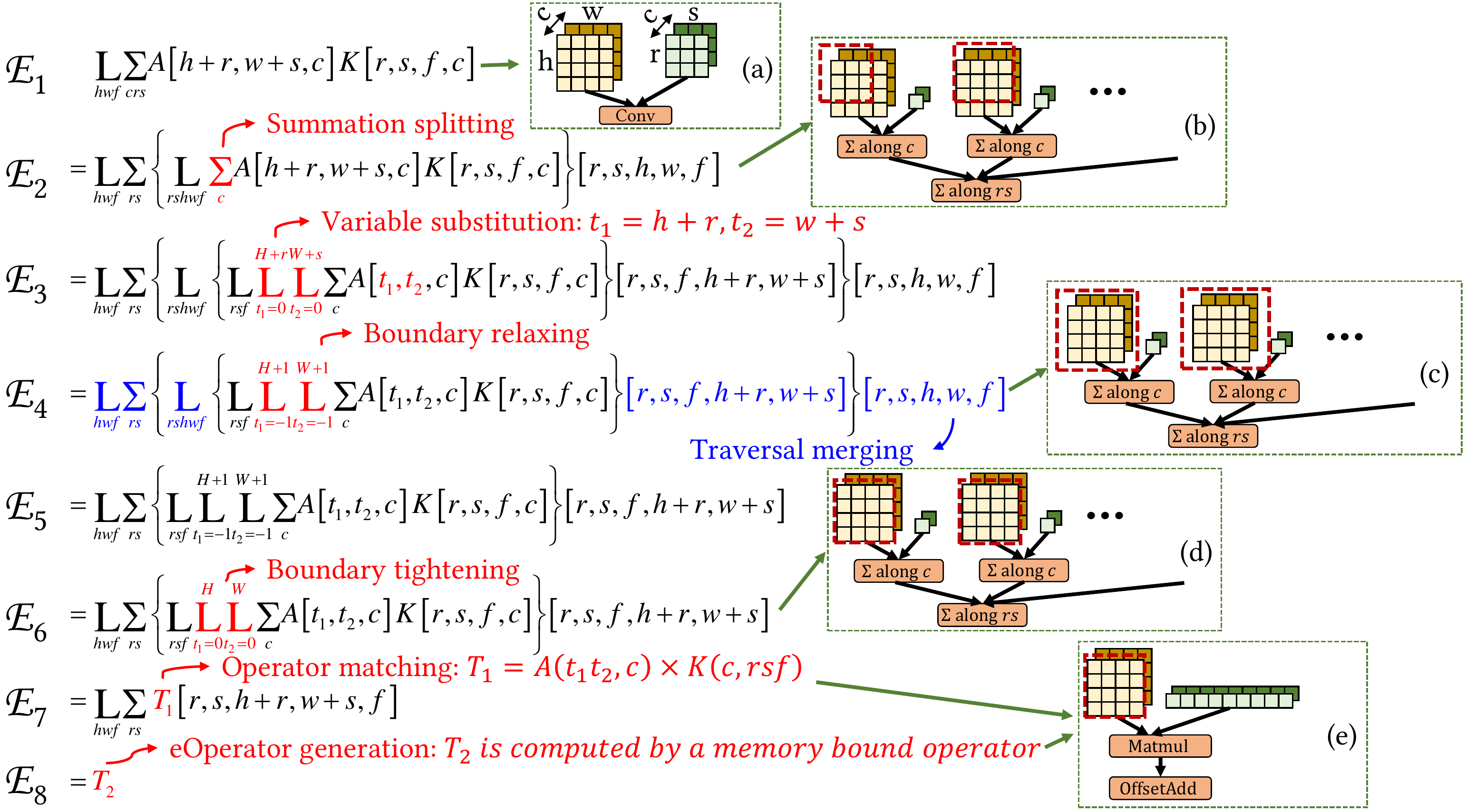}
    \vspace{-1em}
    \caption{The derivation process of the example in \Cref{fig:conv2gemm}, which replaces {\tt Conv} with {\tt Matmul} and \eops}
    \label{fig:conv2gemm-expr}
    \vspace{-1.5em}
\end{figure*}

\subsection{Intra-Expression Derivation}
\label{sec:compute-rule}
Intra-expression derivation rules transform an expression into other functionally equivalent forms, which is essential for constructing a comprehensive search space of expressions for a tensor program.
 \Cref{fig:conv2gemm-expr} shows the intra-expression derivation rules used to discover the optimization in \Cref{fig:conv2gemm}. We now describe these intra-expression derivation rules.
\para{Summation splitting} divides a summation $\sum_{\vec{s}}$ in an expression into two summations $\sum_{\vec{s_1}}$ and $\sum_{\vec{s_2}}$ and instantiates the result of the inner summation by converting it to a \stage:
\vspace{-0.5em}
\begin{align*}
 \opL_{\vec{x}}\sum_{\vec{s_1},\vec{s_2}}f(\mathT\pos{\vec{x},\vec{s_1},\vec{s_2}})\Rightarrow\opL_{\vec{x}}\sum_{\vec{s_1}}\big\{\opL_{\vec{x}\vec{s_1}}\sum_{\vec{s_2}}f(\mathT\pos{\vec{x},\vec{s_1},\vec{s_2}})\big\}[\vec{x},\vec{s_1}]
\end{align*}\\[-1.3em]
where $\mathbf{\tau}$ is a mapping from $(\vec{x}, \vec{s_1}, \vec{s_2})$ to an input position.
\sys divides the iterators of a summation into two disjoint groups, $\vec{s_1}$ and $\vec{s_2}$, which splits the summation into two nested \stages $\m{S}_1, \m{S}_2$, where $\m{S}_1=\opLd_{\vec{x}\vec{s_1}}\sumd_{\vec{s_2}}f(\mathT\pos{\vec{x},\vec{s_1},\vec{s_2}})$ and $\m{S}_2=\opLd_{\vec{x}}\sumd_{\vec{s_1}}\m{S}_1[\vec{x},\vec{s_1}]$.
Note that in summation splitting, \sys converts the result of the inner summation into a \stage, whose output is reused by the outer summation.

To transform a 3$\times$3 convolution to a batch of nine matrix multiplications, as shown in \Cref{fig:conv2gemm-expr}, \sys first transforms the initial expression $\m{E}_1$ to $\m{E}_2$ by splitting the summation $\sum_{crs}$ into two operations $\sum_{rs}$ and $\sum_{c}$, and instantiating the output of the inner summation 
(i.e., \big\{$\opL_{rshwf}\sum_{c}A[h+r,w+s,c]K[r,s,f,c]$\big\}).
The inner \stage only sums along the $c$ dimension; as a result, an intermediate $5$-dimensional tensor is instantiated since the summation along the $r$ and $s$ dimensions is not performed but converted to traversal notations.
The outer \stage computes the remaining summation over the $r$ and $s$ dimensions, which produces a $4$-dimensional tensor.
\Cref{fig:conv2gemm-expr} (a) and (b) depict the computation graphs before and after this derivation.

\para{Variable substitution} substitutes a set of traversal iterators $\opL_{\vec{x}}$ with a new set of iterators $\opL_{\vec{y}}$ by applying a {\em bijective} function $\Phi$ (i.e., $\vec{y}=\Phi(\vec{x})$).\footnote{A mapping function $\Phi$ is {\em bijective} if it is both injective (one-to-one) and surjective (onto).}
This transformation allows the expression to be computed using a different set of traversal iterators. In particular, for an expression $\opL_{\vec{x}}^{\mathbb{X}}f(\mathT[\tau(\vec{x})])$, variable substitution introduces an intermediate \stage that computes $\opL_{\vec{y}}^{\mathbb{Y}}f(\mathT[\tau(\Phi^{-1}(\vec{y}))])$, where $\Phi$ is a bijective function that maps $\mathbb{X}$ to $\mathbb{Y}=\Phi(\mathbb{X})$, and $\Phi^{-1}$ is the reverse function of $\Phi$. Formally,
\aligntop
\begin{align*}
\opL_{\vec{x}}^\mathbb{X}f(\mathT\pos{\vec{x}})\Rightarrow\opL_{\vec{x}}^\mathbb{X}\{\opL_{\vec{y}}^{\Phi(\mathbb{X})}f(\mathT\pos{\Phi^{-1}(\vec{y})})\}[\Phi(\vec{x})]
\alignbottom
\end{align*}
Variable substitution constructs an intermediate \stage with new traversal iterators.
To preserve functional equivalence, the original iterator $\vec{x}$ is used to construct the final result using the output of the intermediate \stage.

In \Cref{fig:conv2gemm-expr}, \sys applies variable substitution to transform the expression from $\m{E}_2$ to $\m{E}_3$, using bijective function: $\Phi(r,s,h,w,f)=(r,s,f,h+r,w+s)$. 
Specifically, $h+r$ and $w+s$ are substituted with $t_1$ and $t_2$ in $\m{E}_3$.
Note that variable substitution does not change the computation graph.

\para{Traversal merging} merges the traversal notations in two \stages into a single \stage using a bijective function $\Phi$:
\aligntop
\begin{align*}
\opL_{\vec{x}}^\mathbb{X}\sum_{\vec{y}}^\mathbb{Y}\{\opL_{\vec{z}}^\mathbb{Z}f(\mathT\pos{\vec{z})}\}[\Phi(\vec{x},\vec{y})]\Rightarrow\opL_{\vec{x}}^\mathbb{X}\sum_{\vec{y}}^\mathbb{Y}f(\mathT\pos{\Phi(\vec{x},\vec{y})})
\alignbottom
\end{align*}
where $\Phi(\mathbb{X}\times \mathbb{Y}) \subseteq \mathbb{Z}$. 

For the example in \Cref{fig:conv2gemm-expr}, \sys applies traversal merging to transform $\m{E}_4$ to $\m{E}_5$.
For this transformation, the outer traversal and summation notations and the inner traversal notation both include five iterators (i.e., $\vec{x}=(h,w,f)$, $\vec{y}=(r,w)$, and $\vec{z}=(r,s,h,w,f)$).
Traversal merging is applied with an identity mapping function $\Phi$ and an indexing function $\tau(r,s,h,w,f)=(r,s,f,h+r,w+s)$.
Traversal merging removes a \stage while preserving the computation graph.

\para{Boundary relaxing and tightening} inspect whether the computation for some output elements can be avoided if these elements are constants for arbitrary inputs.
\sys executes constant propagation on expressions to deal with constant numbers in expressions and padding in tensors.
If a region of output has constant value, \sys converts it into an attribute of tensors to reduce meaningless computation. 
The following formula shows how boundary relaxing and tightening are performed:
\vspace{-1.2em}
\begin{align*}
\opL_{\vec{x}}^\mathbb{X}f(\mathT\pos{\vec{x}}) \Longleftrightarrow \opL_{\vec{x}}^{\mathbb{X}'}f(\mathT\pos{\vec{x}})
\alignbottom
\end{align*}
, where $\mathbb{X} \subset \mathbb{X}'$, and computation on $\mathbb{X}' - \mathbb{X}$ will not be used as the results of the tensor program. 

In the running example in \Cref{fig:conv2gemm-expr}, the formula in $\m{E}_4$ performs boundary relaxing on $t_1$ and $t_2$, transforming their ranges from $[0, H+r)$ and $[0, W+s)$ to $[-1, H+1)$ and $[-1, W+1)$, respectively, as $r$ and $s$ are in range of $[-1, 1]$. 
After boundary relaxing, the computation graph is transformed from \Cref{fig:conv2gemm-expr} (b) to (c).
The formula in $\m{E}_6$ performs boundary tightening on on $t_1$ and $t_2$, transforming their ranges from $[-1, H+1)$ and $[-1, W+1)$ to $[0, H)$ and $[0, W)$, respectively, as the computation taken on $t_1 = -1$, $t_1 = H+1$, $t_2 = -1$ and $t_2 = W+1$ will not be used for the output tensor.
After boundary tightening, the computation graph is transformed from \Cref{fig:conv2gemm-expr} (c) to (d).

\subsection{Expression Instantiation}
\label{sec:expre-contraction}

Expression instantiation rules lower expressions into executable kernels.
After applying these rules, the instantiated scope is replaced with a tensor in the original expression and separated from the following derivation. 
\sys supports two classes of instantiation, operator matching and \eop generation, to leverage existing highly optimized kernels and flexibly generate other kernels.  
In this subsection, we will introduce how these two instantiation rules work.

\begin{table}[]
\caption{\Itab}
\label{tab:itab}
\vspace{-0.5em}
\centering
\small
\begin{tabular}{ccc|cccc}
\toprule
\multicolumn{3}{c|}{Tensors}          & \multicolumn{4}{c}{Ops} \\
\hline
input      & weight     & output     & \texttt{Add} & \texttt{Matmul}  & \texttt{Conv}  & \texttt{G2BMM} \\
\hline
\checkmark & \checkmark & \checkmark & $mn$  & $b$       &       & $bm$    \\
\checkmark &            & \checkmark &     & $m$       & $nhw$   &       \\
           & \checkmark & \checkmark &     & $n$       & $f$     & $w$     \\
\checkmark & \checkmark &            &     & $k$       & $crs$   & $k$     \\
\bottomrule
\end{tabular}
\vspace{-2em}
\end{table}

\subsubsection{Pattern Matching}
\label{sec:rule-pattern-matching}

Mapping a tensor algebra expression to an operator is challenging since an operator can be represented in various expressions with potentially different tensor layouts, tensor shapes, and iterators.
For example, the standard tensor algebra expression of a batched matrix multiplication is shown in Expression~(\ref{eq:bmm}). 
However, although Expression~(\ref{eq:bmm-variant}) is significantly different from Expression~(\ref{eq:bmm}), it can also be directly instantiated as a batched matrix multiplication in math libraries\footnote{Math libraries including BLAS allow for specifying the stride of dimensions to work with different input layouts.}. Similar problems exist for other operators. 
\aligntop
\begin{align}
    \opL_{bmn}\sum_{k}A[b,m,k]B[b,k,n]
    \label{eq:bmm}
\end{align}\\[-2.8em]
\begin{align}
    \opL_{bmn}\sum_{k}C[b,0,m,1+k]D[b-1,b,k,n]
    \label{eq:bmm-variant}
\end{align}\alignbottom

To match a tensor algebra expression to operators, 
\sys uses {\em \itab} to summarize linear algebra properties of expressions.
\Itab supports all operators expressible in tensor algebra expression. \Cref{tab:itab} shows typical operators, including an element-wise operator, matrix multiplication, convolution, and G2BMM~\cite{langer2018band} (general to band matrix multiplication). 
The right part of the table list all iterators of an operator, which are categorized into four groups based on whether the iterators appears in the input, weight, and output tensor. 
\Itab also records the coefficient of iterators in the index of each tensors for operator matching.

Taking the batched matrix multiplication in Expression~(\ref{eq:bmm}) as an example, the iterators $b$, $m$, $n$, and $k$ appear in different combinations of the input, weight, and output tensors.
These variables are assigned to different \emph{iterator groups}, as shown in \Cref{tab:itab}.
To match an expression with a specific operator, we need to match the iterators in this expression to the iterators in the \itab.

The detailed matching process is listed as follows:

\begin{enumerate}[leftmargin=*]
    \item  \textbf{Fill iterators into \itab}: enumerate all the iterators in a given expression, and fill them in into the iterator groups in the \itab according to which tensor it appears in. 
    
    \item \textbf{Match iterator groups}: for each iterator group in \itab, the number of iterators should be the same.
    
    \item \textbf{Match iterators}: if there are multiple iterators in an iterator group, enumerate all one-to-one mapping between variables of the operator and those of the expression. 
    After the variables are mapped, check whether the coefficients of iterators in index match (whether the coefficient of each variable and the correlation between different variables in the expression are consistent with patterns, e.g., $h$ and $r$ should have the same coefficient in the index of input tensor of convolution in Expression~$\m{E}_1$ in \Cref{fig:conv2gemm-expr}). 

\end{enumerate}

For the example in Equation~(\ref{eq:bmm-variant}), it has the same iterator groups with \texttt{Matmul}. There is an unique one-to-one iterator mapping between the expression and \texttt{Matmul} and it matches all iterators.
After the pattern match rule succeeds, \sys replaces the matched scope with a tensor and generates a corresponding \texttt{Matmul} operator in the computation graph.

\subsubsection{\eop Generation}
\label{sec:rule-eop-matching}

\begin{figure}[h]
    \centering
    \vspace{-0.5em}
    \includegraphics[width=\linewidth,keepaspectratio]{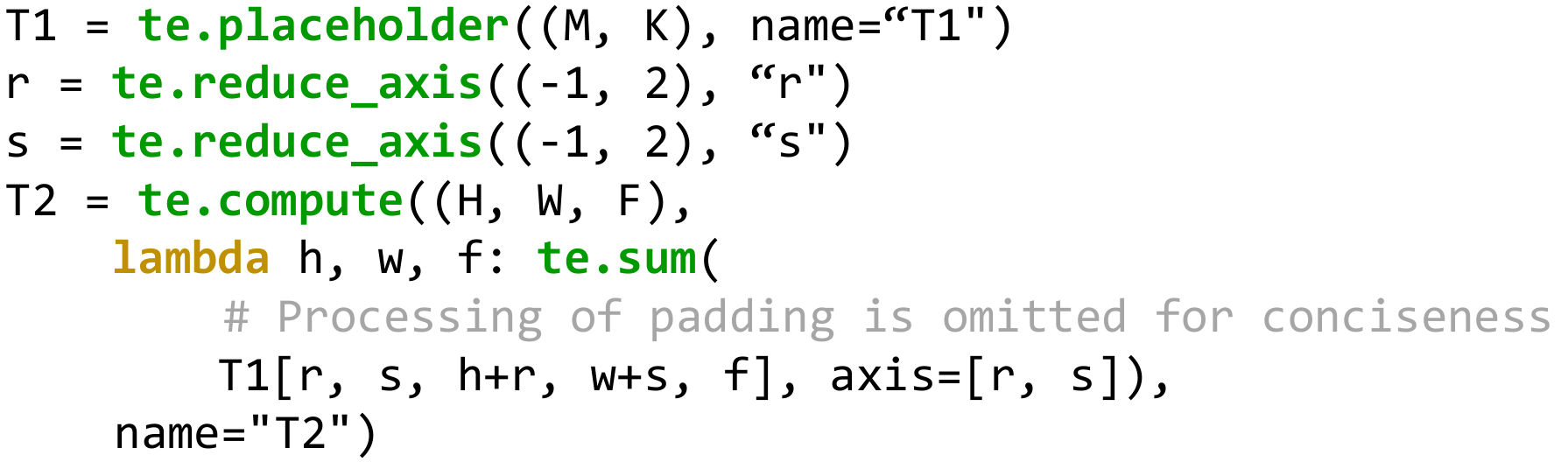}
    \vspace{-2em}
    \caption{Lowering $\m{E}_7$ in \Cref{fig:conv2gemm-expr} to TVM.}
    \label{fig:codegen}
    \vspace{-2em}
\end{figure}

For expressions which cannot match existing operators, \sys converts them into \eops and generates executable kernels based on their tensor algebra expressions.
Since tensor algebra expressions exactly define the computation semantics, 
\sys is able to feed them into existing code generation frameworks, such as TVM, Halide, and Tiramisu~\cite{baghdadi2019tiramisu}.
For example, for the expression in $\m{E}_7$  of \Cref{fig:conv2gemm-expr}, \sys lowers it to TVM by converting its iterator space of traversal notation into a tensor and the computation expression into a lambda function.
\Cref{fig:codegen} shows the result which can be processed by TVM to automatically generate kernel.

\subsubsection{Discussion}

Although \sys can treat arbitrary expressions as \eops to support various optimizations, the code generation process introduces too much overhead making such optimizations impractical.
Considering existing DNN libraries already provide many predefined operators which can outperform generated kernels in most cases, \sys adopts the strategy that mapping the expression to the combination of compute-bound predefined operators and memory-bound \eops.
In this way, \sys can benefit from both the efficiency of existing libraries and the flexibility of auto-generated \eops.
Moreover, tuning memory-bound operators is much easier than tuning compute-bound ones since the former ones often have much easier schedules. 
This guarantees \sys can find optimized transformations within a promising time.
Typically, \sys spends around one minutes to generate a memory-bound \eops.

%% file: search.tex
\section{Derivation-Based Program Optimizations}
In this section, we describe how \sys performs derivation-based transformations to optimize tensor programs in \Cref{sec:inter-expr} and \Cref{sec:derivator}, and related optimization techniques in other subsections.

\subsection{Program-level optimizer}
\label{sec:inter-expr}

\begin{algorithm}[t]
\caption{Program-level optimizer.}
\label{algo:inter-expr}
\begin{algorithmic}[1]
\State {\bf Input:} An input tensor program $\m{P}$
\State {\bf Output:} An optimized tensor program $\m{P}_\text{opt}$
\State
\State $\m{IR} =$ inter-expression rule set
\State $\m{SP}$ = split {$\m{P}$} and translate subprograms into expressions
\State $\m{P}_\text{opt} = \varnothing$
\For {$\er{E} \in \m{SP}$}
\For {$\er{r} \in \m{IR}$}
\State $\m{E}' =$ apply $\er{r}$ to $\er{E}$ \label{line:interRule}
\State $\m{SP} = \m{SP} + \m{E}'$
\EndFor
\EndFor
\For {$\er{E} \in \m{SP}$}
\State $\er{candidates}$ = \Call{HybridDerivation}{$\er{E}$} \label{line:TwoStageDerivation}
\State $\er{best} =$ candidate with best performance in $\er{candidates}$
\State add $\er{best}$ into $\m{P}_\text{opt}$
\EndFor
\State \Call{PostOptimization}{$\m{P}_\text{opt}$}
\State \Return {$\m{P}_\text{opt}$}
\end{algorithmic}
\end{algorithm}

\Cref{algo:inter-expr} shows the workflow of the program-level optimizer.
For an input tensor program, \sys{} first splits it into subprograms using non-linear activation operators as the splitting points. This is because activation operators often do not provide further optimization opportunities other than fusion, as discovered by prior work~\cite{pet}.
For each subprogram, \sys translates it into expressions using the pre-defined expression for each operator. 
Note that a subprogram may contain several operators and thus includes multiple expressions.
For the expressions of each subprogram, \sys{} applies the inter-expressions rules to perform {\em inter-operator} transformations (\Cref{line:interRule}) and then feeds each expression to \sys's hybrid derivation optimizer (see \Cref{sec:derivator}) for generating {\em intra-operator} transformations (\Cref{line:TwoStageDerivation}).
At last, \sys{} merges the transformation with the best performance of each subprogram, performs post-optimizations like fusing the adjacent memory-bound operators, and finally generates an optimized tensor program.

\subsection{Hybrid Derivation Optimizer}
\label{sec:derivator}

\begin{figure}[t]
    \centering
    \includegraphics[width=\linewidth,keepaspectratio]{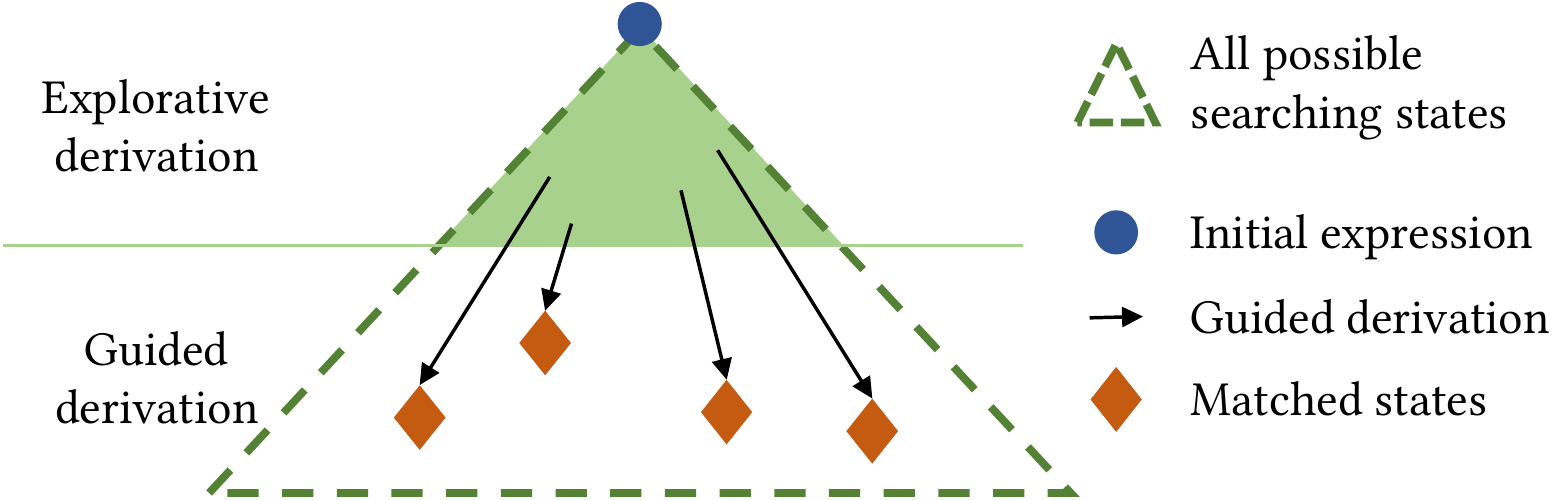}
    \vspace{-1.5em}
    \caption{Hybrid derivation algorithm.}
    \label{fig:search}
    \vspace{-1.5em}
\end{figure}

To explore the vast expression transformation space effectively, \sys leverages a {\em hybrid derivation algorithm} to efficiently discover possible transformations. 
\Cref{fig:search} illustrates the procedure of this hybrid derivation algorithm, which includes an explorative derivation stage and a guided derivation stage.
During the explorative derivation, \sys iteratively applies all derivation rules to current expressions.
However, due to the enormous searching states, only adopting explorative derivation will introduce unacceptable searching overhead.
To reduce searching overhead, \sys adopts guided derivation to match expressions towards target states with the help of \itab. 
As mentioned in \Cref{sec:rule-pattern-matching}, the target states are well optimized operators provided by existing kernel libraries, and \sys can automatically generate corresponding \eops inferred by guided derivation to bridge the gap between current searching state and target searching state.
\Cref{algo:derivation} shows the detailed algorithm of hybrid derivation. 
\begin{algorithm}[t]
\caption{Hybrid derivation algorithm.}
\label{algo:derivation}
\begin{algorithmic}[1]
\State {\bf Input:} An input expression $\m{E}_0$
\State {\bf Output:} A set of tensor program transformations $\m{P}$
\State 
\State $\m{R}$ = intra-expression and instantiation rules set
\State $\m{O}$ = well-optimized operators
\State $\m{P} = \varnothing$  \Comment{Output set}
\State $\m{F} = \varnothing$  \Comment{Fingerprint set}
\State \Call {ExplorativeDerive}{$\m{E}_0$}
\Function{ExplorativeDerive}{$\m{E}$}
\State let $\m{Q}$ be a queue storing expressions and depths
\State $\m{Q}$.queue([$\m{E}_0$, 0])
\While{$\m{Q}$ is not empty}
\State [$\m{E}$, $depth$] = $\m{Q}$.dequeue()
\If {$depth < \er{MaxDepth}$} \label{line:maxdepth}
\State $\er{fp} =$ \Call{FingerPrint}{$\m{E}$}
\If {$\er{fp} \in \m{F}$} \label{line:prune}
    \State \textbf{continue}
\EndIf
\If {$\m{E}$ is a tensor}
    \State $\m{P}=\m{P}\cup\{\m{E}\}$
    \State \textbf{continue}
\EndIf
\State $\m{F}=\m{F}\cup\{\er{fp}\}$
\For {$\er{r} \in \m{R}$} \Comment{Select and apply a rule}
    \State $\m{Q}$.queue([$\er{r}(\m{E})$, $depth+1$])
\EndFor
\EndIf
\For{$\m{T} \in \m{O}$}
    \State \Call{GuidedDerive}{$\m{E}$, $\m{T}$} 
\EndFor
\EndWhile
\EndFunction
\State

\Function{GuidedDerive}{$\m{E}$, $\m{T}$}
\If {$\m{E}$ is a tensor}
    \State $\m{P}=\m{P}\cup\{\m{E}\}$ \label{line:single-tensor}
    \State \Return
\EndIf
\For {$\er{r} \in$ \Call{SelectRules}{$\m{E}$,$\m{T}$}}
	\State \Call{GuidedDerive}{$\er{r}(\m{E})$, $\m{T}$} \label{line:derictional}
\EndFor
\EndFunction
\end{algorithmic}
\end{algorithm}

\para{Explorative derivation}
During the explorative derivation, \sys tries to apply the derivation rules on the input expression for certain rounds (called $depth$), to explore different functionally-equivalent expressions. 
The search is controlled by a hyperparameter $\er{MaxDepth}$, which determines the maximum depth of derivation steps during explorative derivation (\Cref{line:maxdepth}). 
To eliminate redundancy, \sys generates a fingerprint (\Cref{sec:fingerprint}) for each discovered expression.
If there is a fingerprint conflict, 
\sys prunes this expression to avoid deriving duplicate expressions (\Cref{line:prune}). 

\para{Guided derivation}
\sys leverages the guided derivation to derive expressions toward well-optimized operators provided by vendor or other libraries. 
For each state in explorative derivation, \sys selects well-optimized operators $\m{T}$ as target for guided derivation.
When the expression is derived to a single tensor (\Cref{line:single-tensor}), 
all computation in the expression are replaced with operators or \eops, which means
an equivalent tensor program is discovered for the input expression.
\sys adds the derived computation graph to the output set.
Otherwise, \sys continues to leverage guided derivation to derive the current expressions toward the pattern of target operators (\Cref{line:derictional}).

\sys decides derivation rules by analyzing iterator mapping table mismatches between current expressions and target operators. 
For the iterator group mismatch, which means more or less iterators exist in a iterator group, \sys merges or splits iterators by applying the variable substitution rule on them. 
While for the iterator mismatches, \sys transforms data layouts to satisfy requirements of target operators.
Since data layout transformations are achieved by variable substitution on tensor algebra expressions,  
\sys constructs the iterator mapping function $\Phi$ by looking up \itab and the constraints of target operators.
For each tensor $\textbf{T}$ with index variables $\vec{x}$, \sys constructs a new tensor, $\textbf{T}'$ with index variables $\vec{x}'$ where $\textbf{T}'[\vec{x'}]=\textbf{T}[\Phi({\vec{x}})]$.

For example, the expression in the inner \stage of $\m{E}_6$ in \Cref{fig:conv2gemm-expr} is mapped to a {\tt Matmul} operator using the \itab.
By comparing the iterators in the expression and the iterators of a \texttt{Matmul} operator in \Cref{tab:itab}, we obtain the following \textit{iterator mapping}:

\aligntop
$$
    t_1, t_2 \rightarrow m; r, s, f \rightarrow n; c \rightarrow k
$$
\alignbottom

Thus, to perform this mapping, iterators $t_1$ and $t_2$ should be fused to a single variable $m$, so do $r$, $s$ and $f$ to $n$.
After mapping iterators, \sys can infer the shape of each tensor from \itab  and construct new tensors from existing tensors.
The input tensor $\textbf{A}'$ and weight tensor $\textbf{K}'$ for {\tt Matmul} are constructed by the following equation:
\vspace{-0.4em}
\begin{align}
    \textbf{A}'[m,k]=\textbf{A}'[t_1 \times W+t_2,c]=\textbf{A}[t_1,t_2,c]
    \label{eq:layout-A}
\end{align}
\alignbottom
\aligntop
\begin{align}
    \textbf{K}'[k,n]=\textbf{K}'[c,r \times S \times F+s \times F+f]=\textbf{K}[r,s,f,c]
    \label{eq:layout-K}
\end{align}
\\[-1.5em]
, where the mapping functions $(m,k)=\Phi_{A}(t_1,t_2,c)=(t_1 \times W+t_2,c)$, $(k,n)=\Phi_{K}(r,s,f,c)=(c,r \times S \times F+s \times F+f)$, and $W$, $S$, and $F$ are the range size of iterators $w$, $s$ and $f$.
\sys automatically generates required tensor layout transformations by instantiating \Cref{eq:layout-A} and (\ref{eq:layout-K}).

\subsection{Redundancy Pruning}
\label{sec:fingerprint}
Given an input tensor algebra expression and derivation rules, \sys's optimizer considers all possible combinations of these rules up to a fixed size.
However, directly enumerating all possible combinations of rules results in significant redundancy since applying different sequences of derivation rules may reach the same expression.
For example, applying the variable substitution rules twice with mapping function $\Phi$ and $\Psi$ generates the following expression:

\begin{align*}
\opL_{\vec{x}}f(\vec{x}) &= \opL_{\vec{x}}\{\opL_{\vec{y}}f(\Phi^{-1}(\vec{y}))\}[\Phi(\vec{x})] \\
&=\opL_{\vec{x}}\{\opL_{\vec{y}}\{\cblack{\opL_{\vec{z}}f(\Phi^{-1}(\Psi^{-1}(\vec{z})))}\}[\Psi(\vec{y})]\}[\Phi(\vec{x})]
\end{align*}
\alignbottom

In the case $\Psi=\Phi^{-1}$, the most inner \stage in the second line above can be transformed to 

\vspace{-1.6em}
\begin{align*}
    \opL_{\vec{z}}f(\Phi^{-1}(\Psi^{-1}(\vec{z})))=\opL_{\vec{z}}f(\Phi^{-1}(\Phi(\vec{z})))=\opL_{\vec{z}}f(\vec{z})
\end{align*}\\[-2.5em]

Note that $\opL_{\vec{z}}f(\vec{z})$ is identical to the original expression $\opLd_{\vec{x}}f(\vec{x})$.
Therefore, we do not need to apply additional derivation rules on this \stage.

\sys uses a fingerprint technique to detect redundant \stages. The {\em fingerprint} is a hash of the expression. It is capable to distinguish the following redundant expressions:
\begin{itemize}[leftmargin=*]
    \item \textbf{Iterator renaming}: iterators should be distinguished by their iterator space, instead of their names, e.g., expressions $\opLd_{x=0}^{N}\opLd_{y=0}^{M}f(x,y)$ and $\opLd_{y=0}^{N}\opLd_{z=0}^{M}f(y,z)$ are equivalent, and $(x,y)$ in the former one should be mapped to $(y,z)$ in the latter one.
    \item \textbf{Summation reordering}: summations can be reordered, e.g., $\sumd_{\vec{x}}\sumd_{\vec{y}}f(\vec{x},\vec{y})$ is equivalent with $\sumd_{\vec{y}}\sumd_{\vec{x}}f(\vec{x},\vec{y})$. Note that traversal reordering is not equivalent transformation since it indicates layout transformation.
    \item \textbf{Operands reordering}: operands of commutative binary operations can be reordered, e.g., $\opLd_{\vec{x}}(\mathbf{T_1}[\vec{x}]+\mathbf{T_2}[\vec{x}])$ equals to $\opLd_{\vec{x}}(\mathbf{T_2}[\vec{x}]+\mathbf{T_1}[\vec{x}])$. Operands reordering should be applied for both iterator computation and tensor computation.
    \item \textbf{Tensor renaming}: tensors introduced by different scopes may has the same value.
\end{itemize}

To satisfy the above features, \sys adopts the following design.
For a traversal iterator, we use its iterator space and its order relative to all the traversal notations in the current \stage as its fingerprint. 
The order that appeared in the fingerprint differentiates traversal iterators with the same iterator spaces but the different locations in the traversal notations. 
While for summation iterator, we only use its iterator space as its fingerprint.
Thus expressions under summation reordering have the same fingerprint.
To satisfy operands reordering, we use the operation name and commutative hash for commutative operations, such as addition and non-commutative hash for other operators.
The fingerprint of tensors depends on their source.
For input tensors, \sys calculates the fingerprint by hashing the tensor name.
For tensors generated by \stages, their fingerprints are defined by the expressions which generate them.

%% file: implementation.tex
\subsection{Post-Processing}

\begin{figure}[t]
    \centering
    \includegraphics[width=\linewidth,keepaspectratio]{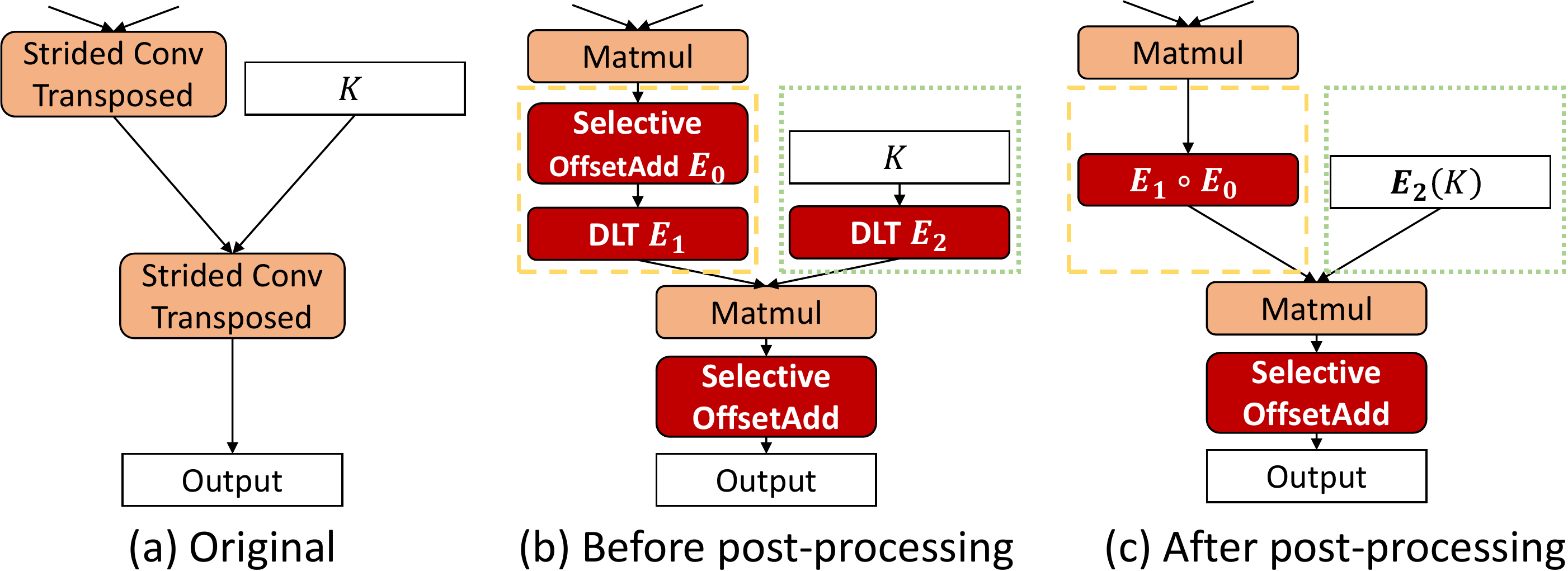}
    \vspace{-2em}
    \caption{Detailed post-processing for InfoGAN. Red blocks represent \eops. DLT means data layout transformation.}
    \label{fig:post-opt}
    \vspace{-1.5em}
\end{figure}

\para{\eop fusion}
\sys generates \eops to facilitate performant transformations.
However, although \eops contain relatively small computation, too many \eops impact performance as well.
To reduce this overhead, \sys fuses consecutive \eops using inter-expression derivations. 
For example,  \Cref{fig:post-opt} shows how \sys optimizes the InfoGAN model. 
In the dashed box of \Cref{fig:post-opt}(b), two \eops are generated and fused into one kernel by the expression fusion rule.

\para{Identity \eop elimination}
\eops produced in derivation can generate identical outputs with inputs, which is called an identity \eop.
For example, although \Cref{eq:layout-A} transforms the tensor data layout, but the underlying data remains the same and can be eliminated. 
To identify identity \eops, \sys squashes input and output tensors to one-dimensional tensor and checks whether the expression is an identical mapping from input to output. 

\para{Compile-time expression evaluation}
For expressions only taking weights as input tensors, \sys computes them at compile time. 
For example, in the dotted box of \Cref{fig:post-opt}, data layout transformation $E_2$ on weight K is computed during post postprocess optimization instead of runtime.

%% file: evaluation.tex
\section{Evaluation}

\subsection{Experimental Setup}

\para{Platform}
We evaluate \sys on a server with dual Intel Xeon CPU E5-2680 v4 CPUs, an A100 40GB PCIe GPU, and a V100 32GB PCIe GPU.
All experiments use CUDA 11.0.2, cuBLAS 11.1.0, and cuDNN 8.0.3.

\para{Workloads} 
We evaluate \sys on seven deep learning models, which span various fields and cover both classic models such as ResNet-18 and emerging language models like LongFormer.
InfoGAN~\cite{chen2016infogan} is a popular generative adversarial network learning disentangled representations from objects.
DCGAN~\cite{radford2015unsupervised_dcgan} leverages deep convolution structures to get hierarchical representations.
SRCNN~\cite{dong2014srcnn} is the first convolutional neural network for image super-resolution.
GCN~\cite{peng2017large_gcn} is an image semantic segmentation model named global convolutional network.
ResNet-18~\cite{resnet} is a famous image classification network.
CSRNet~\cite{li2018csrnet} adopts dilated convolution for congested scene analysis.
LongFormer~\cite{beltagy2020longformer} is an improved Transformer architecture network dealing with long-sequence language processing with dilated attention.

\begin{figure*}[t]
    \centering
    \includegraphics[width=\linewidth,keepaspectratio]{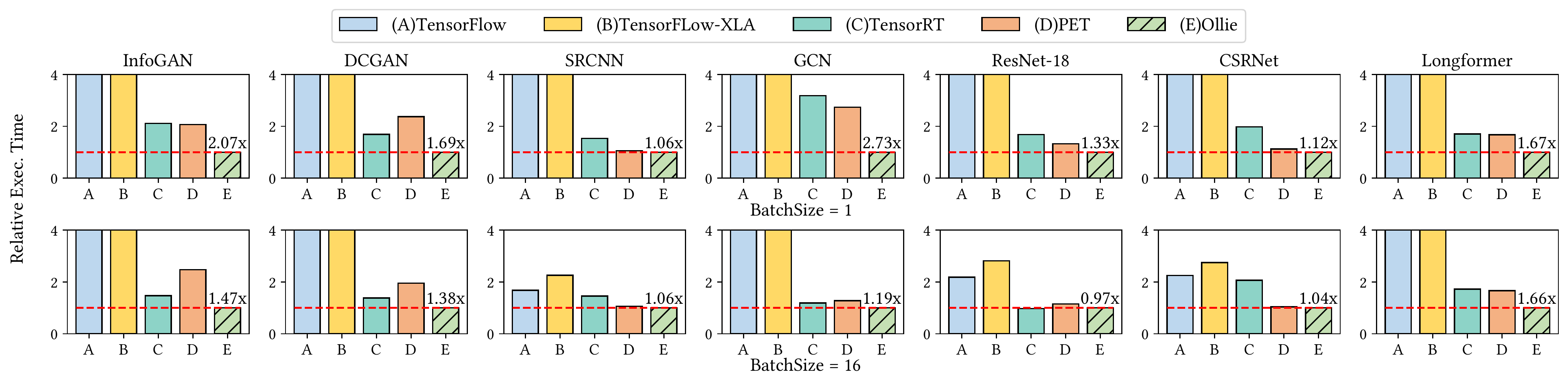}
    \vspace{-2.5em}
    \caption{End to end performance on A100 with batch size 1 and 16.}
    \label{fig:eval-e2e-a100}
    \vspace{-1em}
\end{figure*}

\begin{figure*}[t]
    \centering
    \includegraphics[width=\linewidth,keepaspectratio]{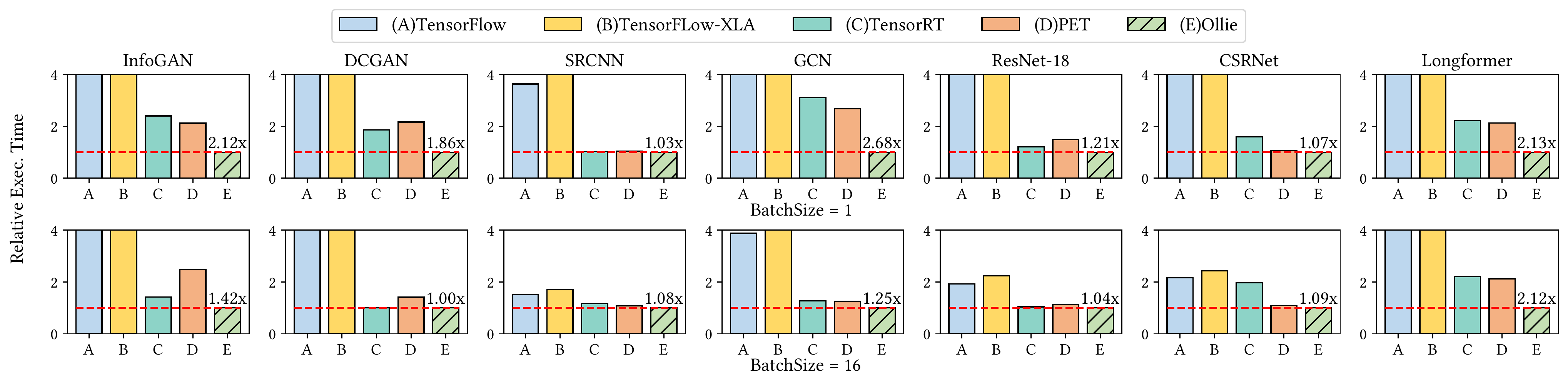}
    \vspace{-2.5em}
    \caption{End to end performance on V100 with batch size 1 and 16.}
    \vspace{-1.5em}
    \label{fig:eval-e2e-v100}
\end{figure*}

\subsection{End-to-End Performance}
\label{subsec:eval_e2e_perf}

We first compare the end-to-end inference time of \sys with popular DNN frameworks, including TensorFlow~\cite{abadi2016tensorflow} 2.4, TensorFlow XLA~\cite{tensorflow_xla}, TensorRT~\cite{tensorrt} 8.2, and PET~\cite{pet}.
All frameworks use the same version cuBLAS and cuDNN libraries as backend to fairly show the improvement from \sys.
For \sys, we set the maximum searching depth to 7.
For the Longformer case, we implement the new attention operator \texttt{G2BMM} by einsum in TensorFlow and provide an auto-tuned kernel for TensorRT, PET, and \sys, since it is not included in the cuBLAS and cuDNN libraries.
\Cref{fig:eval-e2e-a100} and \Cref{fig:eval-e2e-v100} shows the end-to-end execution time on A100 and V100 respectively.
Detailed operator performance analysis will be shown in \Cref{subsec:eval_op_ansor}.

\sys achieves speedups up to 2.73$\times$ on A100 and 2.68$\times$ on V100.
For classical model ResNet-18, \sys still achieves speedups up to 1.33$\times$.
\sys discovers new optimizations which cannot be found by previous frameworks, such as transforming convolution to \texttt{Matmul} and \texttt{OffsetAdd} (\Cref{fig:conv2gemm}).
PET finds many effective optimization and outperforms TensorRT on SRCNN, GCN, and CSRNet.
However, \sys still achieves speedups over PET since more performant transformations for the same subprograms are discovered.
For CSRNet, one of the typical optimizing cases of PET, \sys transforms the dilated convolution into non-dilated convolution by expression derivation.
This optimization is similar to PET, indicating that \sys can contain the optimizations in PET, and discovers further optimizations.
We will discuss typical optimizations in \Cref{sec:Opt_analysis}.

\begin{table}[t]
\caption{\newcnt[C1.1]{
Performance studies on the operators in~\Cref{sec:Opt_analysis}.
{\tt IGEMM}, {\tt FFT}, and {\tt WINO} in the Algo column refer to implicit GEMM, Fast Fourier Transform, and Winograd convolution algorithms, respectively. 
The DRAM and L2 columns show the amount of memory access.
The input shapes for the following operators are [1, 512, 7, 7], [16, 256, 2, 2], [16, 32, 224, 224] and [8, 10000, 64], respectively. 
The first three cases are based on cuDNN and cuBLAS, and the last one case uses kernel generated from Ansor.
}
}
\vspace{-1em}
\label{tab:conv-perf-detail}
\scriptsize
\centering
\resizebox{0.99\linewidth}{!}{%
\begin{tabular}{cccccc}
\toprule
    & & \textbf{Algo}  & \parbox{2em}{\centering\textbf{Time\\(ms)}} & \parbox{4em}{\centering\textbf{DRAM\\(MB)}} & \parbox{4em}{\centering\textbf{L2\\(MB)}}    \\

\midrule
{\tt Conv3x3} & {\textbf{Original}} & {\tt WINO} & $0.126$ & $56.7$ & $70.6$  \\
\Cref{fig:conv2gemm} & {\textbf{Optimized}} & {\tt N/A}   & $0.046$ & $10.5$ & $27.5$  \\

\midrule
{\tt ConvTranspose} & {\textbf{Original}} & {\tt IGEMM} & $0.136$ & $7.74$ & $122$  \\
\Cref{fig:tconv} & {\textbf{Optimized}} & {\tt N/A}   & $0.032$ & $8.07$ & $27.8$   \\

\midrule
{\tt Conv5x5} & {\textbf{Original}} & {\tt FFT} & $0.854$ & $547$ & $579$  \\
 & {\textbf{Optimized}} & {\tt WINO}   & $0.528$ & $368$ & $499$   \\

\midrule
{\tt G2BMM}& {\textbf{Original}} & {\tt N/A} & $20.3$ & $129$ & $52690$  \\
 & {\textbf{Optimized}} & {\tt N/A}   & $2.59$ & $41.1$ & $1400$  \\

\bottomrule
\end{tabular}%
}
\vspace{-2em}
\end{table}

\subsection{Optimization Analysis}
\label{sec:Opt_analysis}

\begin{figure}[t]
    \centering
    \includegraphics[width=\linewidth,keepaspectratio]{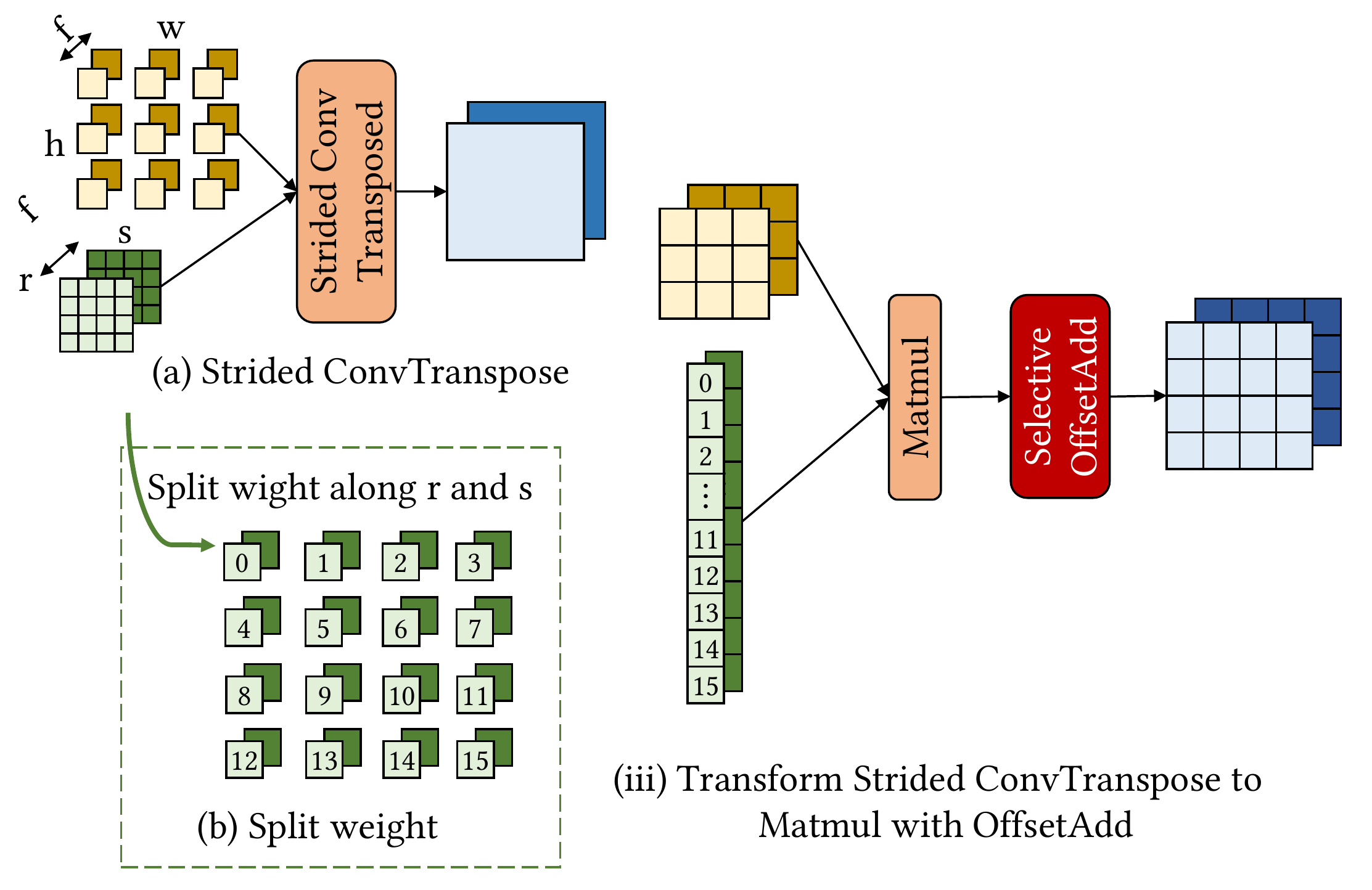}
    \vspace{-2em}
    \caption{Strided \texttt{ConvTransposed} to \texttt{Matmul} in InfoGAN}
    \vspace{-1em}
    \label{fig:tconv}
\end{figure}

To find how \sys optimizes models, we illustrate typical optimizations discovered by \sys and conduct performance analysis.
Their configurations are shown in \Cref{tab:conv-perf-detail}.

For the \texttt{Conv3x3} in ResNet-18, \sys transforms it to \texttt{Matmul} and an \eop \texttt{OffsetAdd} according to \Cref{fig:conv2gemm}.
\texttt{OffsetAdd} will be fused with following element-wise operators.
The profiling data in \Cref{tab:conv-perf-detail} shows that less read and write are required and \sys achieves a 3.65$\times$ speedup against cuDNN which selecets Winograd algorithm.

\sys transforms the \texttt{ConvTranspose} in InfoGAN to \texttt{Matmul} and an \eop.
As shown in \Cref{fig:tconv}, \sys directly applies \texttt{Matmul} to the kernel and input tensor without padding.
Since the \texttt{ConvTranspose} is strided, its input tensor is padded among adjacent elements.
For one output element, only part of the outputs of \texttt{Matmul} are required to be summed up. 
\sys automatically performs transformation and produces a selective addition on the outputs of \texttt{Matmul} by expression derivation.
In this transformation, a direct computation of \texttt{GEMM} can avoid the redundant computation on \texttt{ConvTranspose} and significantly reduce the L2 access which makes the main contribution of the performance optimization.

\subsection{Collaboration with Different Backends}
\label{subsec:eval_op_ansor}

\begin{figure}[t]
    \centering
    \includegraphics[width=0.9\linewidth,keepaspectratio]{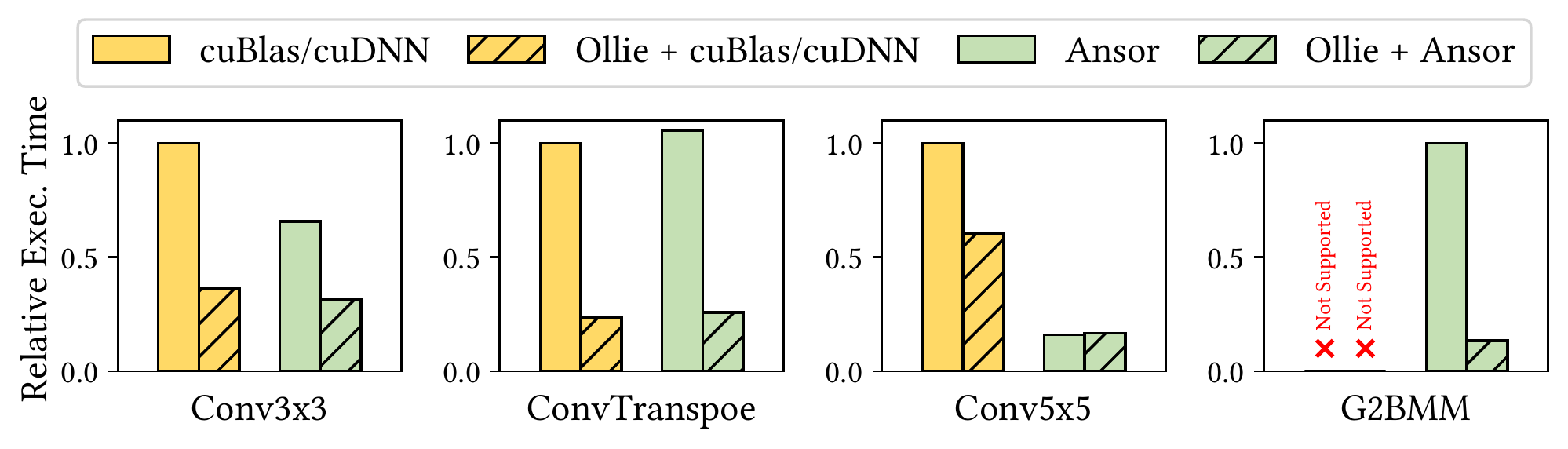}
    \vspace{-1em}
    \caption{Operator performance before and after optimization (opt) on the math libraries and code generation framework Ansor. The input settings are shown in \Cref{tab:conv-perf-detail}.}
    \vspace{-1.5em}
    \label{fig:eval-operator}
\end{figure}

Since \sys searches expression space, it can cooperate with different backends, including math libraries and schedule-based code generation frameworks.
To show the improvements of \sys on these backends, we evaluate \sys with cuBLAS/cuDNN and TVM~\cite{tvm} with its auto-scheduler Ansor~\cite{tvm_auto_tuner}.
The evaluation is carried out on the same transformations illustrated in \Cref{sec:Opt_analysis}.

\Cref{fig:eval-operator} shows \sys is beneficial for different backends.
For \texttt{Conv3x3} in ResNet-18 and \texttt{ConvTranspose} in InfoGAN, transforming it to \texttt{Matmul} and an \eop has significant speedup over both cuDNN and Ansor. 
The transformation from \texttt{Conv5x5} to \texttt{Conv3x3} in SRCNN is beneficial for cuDNN. Such transformation do not have performance improvement on Ansor since Ansor directly tunes original computation, and it cannot adopt efficient algorithms such as Winograd~\cite{lavin2016fast} which is preferred for \texttt{Conv3x3}.
For {\tt G2BMM} operator in Longformer, which is not a pre-defined operator in math libraries, \sys is able to transform it from a dilated form to non-dilated form, which brings a speedup of 7.49$\times$ since less non-contiguous memory access happens.

\subsection{Analysis on Automated Derivation}

\begin{figure}[t]
    \centering
    \includegraphics[width=\linewidth,keepaspectratio]{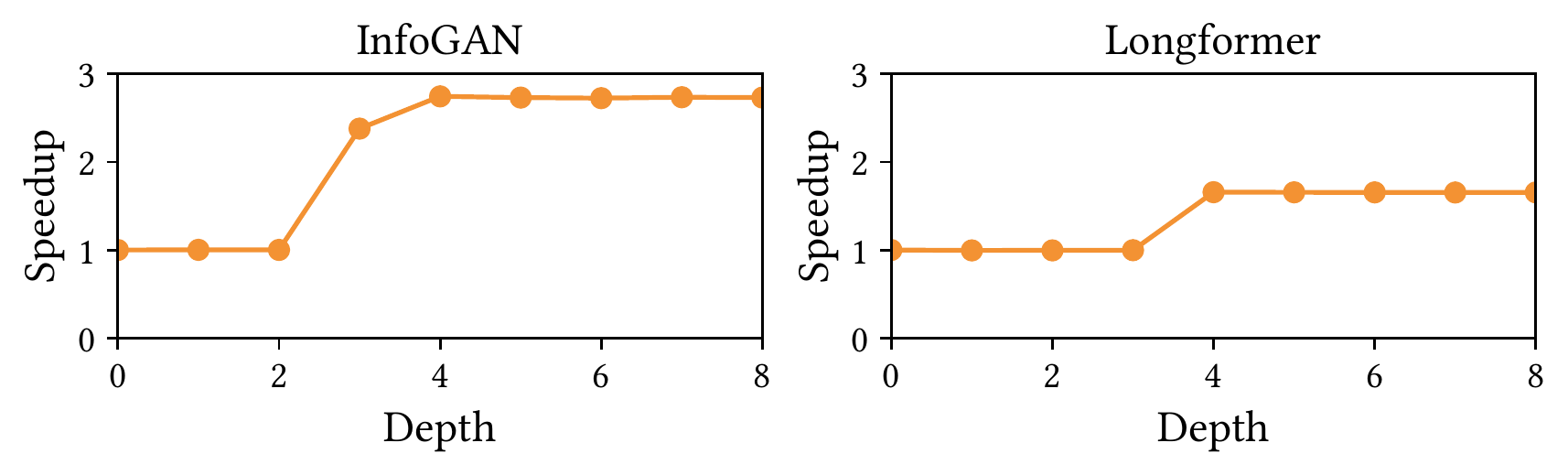}
    \vspace{-2.5em}
    \caption{Speedup under different maximum search depths}
    \vspace{-1em}
    \label{fig:eval-search}
\end{figure}

The search space of \sys is determined by several heuristic parameters, such as the maximum searching depth in the automated derivation algorithm (\Cref{algo:derivation}). 
The maximum depth specifies the largest steps of derivation applied to an expression.
A larger searching depth enables more potential optimizations but larger searching overhead. 
In \Cref{fig:eval-search}, we analyze the speedup \sys can achieve with different maximum searching depth on InfoGAN and Longformer. 
On InfoGAN, \sys has improvement when searching step increases from 2 to 4, as new transformations are explored with a deeper search. 
While for Longformer, the major performance speedup comes from the transformation found in a 4-step derivation. 
In conclusion, the key takeaway is that \sys can achieve most of the performance improvement at moderate depth.

To evaluate the proposed techniques for derivation, we evaluate the searching process on the four cases in \Cref{tab:conv-perf-detail} with and without guided derivation and expression fingerprint.

\begin{figure}[t]
    \centering
    \includegraphics[width=\linewidth,keepaspectratio]{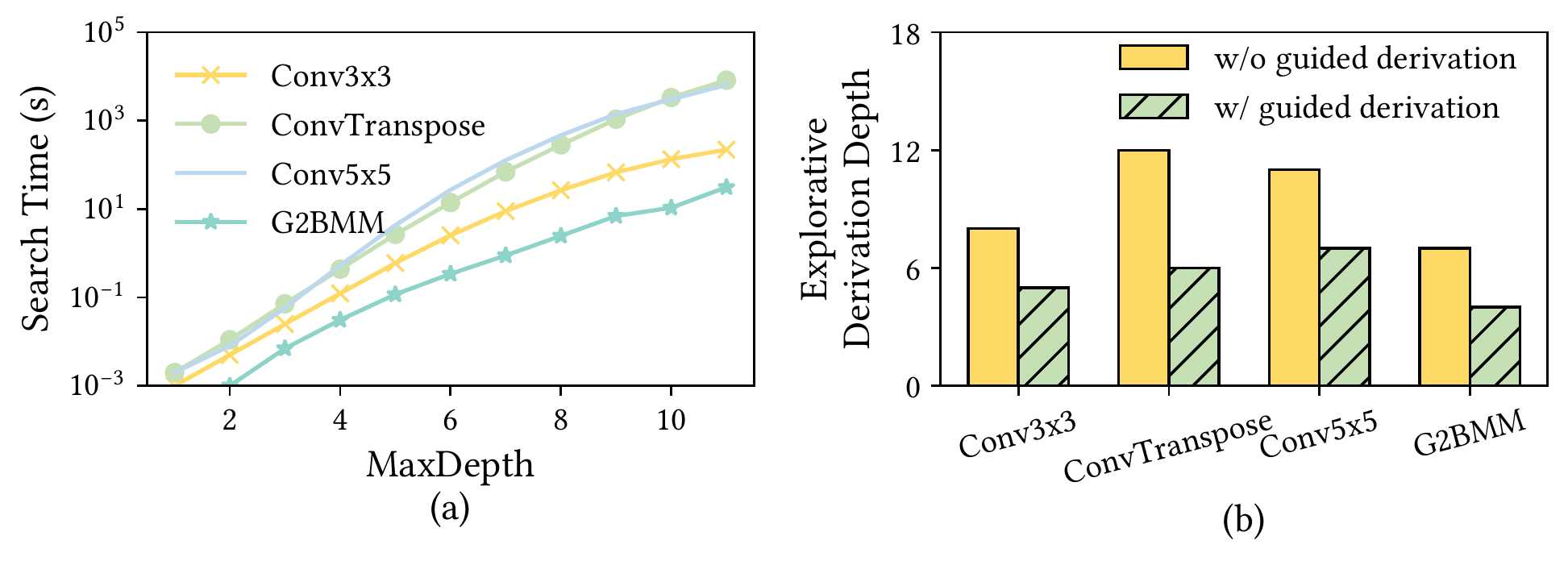}
    \vspace{-2.5em}
    \caption{(a) Search time with different $MaxDepth$. (b) The number of explorative derivation steps with and without guided derivation.}
    \label{fig:eval-guideSearch}
    \vspace{-1.5em}
\end{figure}

\para{Hybrid derivation (\Cref{sec:derivator})} provides a deterministic derivation direction to avoid enormous search overhead.
As shown in \Cref{fig:eval-guideSearch}(a), the search time grows exponentially with the maximum depth of explorative derivation (i.e., $MaxDepth$ in \Cref{algo:derivation}). 
\sys adopts guided derivation to replace explorative derivation.
\Cref{fig:eval-guideSearch}(b) shows the number of applied explorative and guided derivations in these cases.

In the case of ConvTranspose, the explorative derivation requires a search with $MaxDepth=12$ to discover the target expression. But with guided derivation, \sys only requires a search with $MaxDepth=6$, which means that matching a vendor-provided operator needs a $6$-step ($12 - 6$) searching and guided derivation can reduce this unnecessary searching. Thus, this optimization leads to a significant reduction of the searching time by more than $99.0\%$.

\begin{figure}[t]
    \centering
    \includegraphics[width=\linewidth,keepaspectratio]{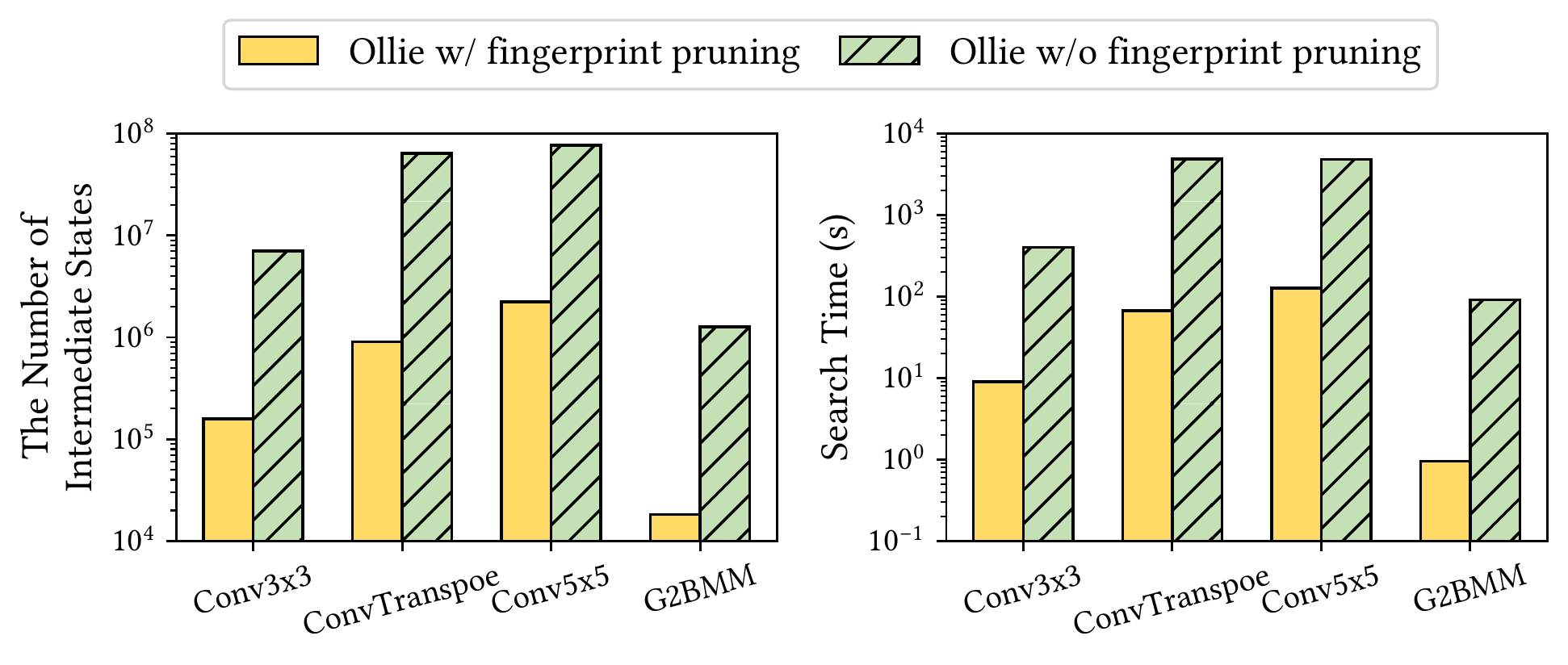}
    \vspace{-2.5em}
    \caption{Ablation study of expression fingerprint pruning}
    \label{fig:eval-ablation}
    \vspace{-1.8em}
\end{figure}

\para{Expression fingerprint (\Cref{sec:fingerprint})} prunes redundant searching states. 
\Cref{fig:eval-ablation} shows the intermediate states and searching time with and without the fingerprint mechanism. During the enumerating derivation, fingerprint effectively prunes $98.0\%$ of intermediate states by recognizing and reducing duplicate expressions and reduces $98.2\%$ of searching time on average.

With the guided derivation and expression fingerprint, \sys finishes searching within two minutes for most subprograms.
The searching time of \sys is therefore on par with other search-based DNN optimization frameworks, such as TASO, PET, and Ansor.
\sys is able to be deployed in real production environment since the searching cost is one-off for a model and brings persistent benefits.

%% file: related.tex
\section{Related Work}

\para{Rule-Based Approaches.} Rule-based optimizations are widely used in different tensor programming frameworks.
Based on TensorFlow~\cite{tensorflow2015-whitepaper}, Grappler~\cite{grappler} performs rule-based optimizations like constant folding and layout optimization.
TensorFlow XLA~\cite{tensorflow_xla}, TensorRT~\cite{tensorrt}, and DNNFusion~\cite{dnnfusion} similarly perform neural network optimizations.
Although rule-based optimizations can efficiently optimize computation graphs, summarizing and implementing such rules require a large amount of human efforts and expertise.

\para{Superoptimization-Based Approaches.}
TASO~\cite{taso} and PET~\cite{pet} try to save human effort in DNN optimization by searching optimized transformations given a set of operators. 
TASO adopts formal verification techniques to assure the equivalence of transformations.
PET further introduces inequivalent transformations and correction mechanism to find more optimizations.
Compared with these frameworks, \sys extends the search space from POR expressions to general expressions by tensor algebra expression derivation. 

\para{Schedule-Based Optimizations.} 
Halide~\cite{DBLP:conf/pldi/Ragan-KelleyBAPDA13}  decouples a program into computation and schedule and performs schedule space transformations.
This idea is widely adopted by many tensor optimization frameworks including TVM~\cite{tvm}, FlexTensor\cite{flextensor}, and Ansor~\cite{ansor}. 
Orthogonal to schedule-based optimizers, \sys focuses on expression space and designs the corresponding expression derivation rules, such as operator matching, to facilitate these new transformations.

%% file: conclusion.tex
\section{Conclusion}

We propose \sys, the first derivation-based tensor program framework that enables expression-level optimizations,
which extends the search space of tensor programs from predefined operator representable expressions to general expressions.
Evaluation results show that \sys outperforms state-of-the-art frameworks by up to $2.73\times$ on NVIDIA A100 and up to $2.68\times$ on NVIDIA V100.